\documentclass[lettersize,journal]{IEEEtran}
\usepackage{amsmath,amsfonts}
\usepackage{algorithmic}
\usepackage{algorithm}
\usepackage{array}
\usepackage[caption=false,font=footnotesize,labelfont=rm,textfont=rm]{subfig}
\usepackage{textcomp}
\usepackage{stfloats}
\usepackage{url}
\usepackage{verbatim}
\usepackage{graphicx}
\usepackage{cite}
\usepackage{color}
\usepackage{colortbl}
\usepackage{makecell}
\usepackage{threeparttable}
\usepackage{booktabs} 
\usepackage{subfloat}
\usepackage{multirow}
\usepackage{amsmath}
\usepackage{amsthm}
\usepackage[hang,flushmargin]{footmisc}
\newtheorem{theorem}{Theorem}[section]

\theoremstyle{definition}
\newtheorem{definition}[theorem]{Definition}

\theoremstyle{remark}

\allowdisplaybreaks

\begin{document}

\title{Similarity-based Label Inference Attack against Training and Inference of Split Learning}
\author{Junlin Liu, Xinchen Lyu, Qimei Cui, and Xiaofeng Tao%
\thanks{This work was supported in part by the National Science Foundation of China under Grant 62371059, in part by the Fundamental Research Funds for the Central Universities under Grant 2242022k60006, and in part by the major key project of PCL under Grant PCL2021A15. (Corresponding Author: Xinchen Lyu).}
\thanks{J. Liu is with the National Engineering Research Center for Mobile Network Technologies, Beijing University of Posts and Telecommunications, Beijing 100876, China. (Email: zero10walker@gmail.com)}%
\thanks{X. Lyu, Q. Cui, and X. Tao are with the National Engineering Research Center for Mobile Network Technologies, Beijing University of Posts and Telecommunications, Beijing 100876, China, and the Department of Broadband Communication, Pengcheng Laboratory, Shenzhen 518055, China. (Email: \{lvxinchen,cuiqimei,taoxf\}@bupt.edu.cn)}%
}

\IEEEpubid{}

\maketitle

\begin{abstract}
Split learning is a promising paradigm for privacy-preserving distributed learning.
The learning model can be cut into multiple portions to be collaboratively trained at the participants by exchanging only the intermediate results at the cut layer.
Understanding the security performance of split learning is critical for many privacy-sensitive applications.
This paper shows that the exchanged intermediate results, including the smashed data (i.e., extracted features from the raw data) and gradients during training and inference of split learning, can already reveal the private labels.
We mathematically analyze the potential label leakages and propose the cosine and Euclidean similarity measurements for gradients and smashed data, respectively.
Then, the two similarity measurements are shown to be unified in Euclidean space.
Based on the similarity metric, we design three label inference attacks to efficiently recover the private labels during both the training and inference phases.
Experimental results validate that the proposed approaches can achieve close to $100\%$ accuracy of label attacks.
The proposed attack can still achieve accurate predictions against various state-of-the-art defense mechanisms, including DP-SGD, label differential privacy, gradient compression, and Marvell.
\end{abstract}


\section{Introduction}
\label{sec:intro}
Artificial intelligence (AI) has opened up new opportunities in various areas of speech recognition~\cite{Hinton12, Alam20}, computer vision~\cite{Krizhevsky17, Nie18}, and manufacturing~\cite{wang2018deep, essien2020deep}.
According to Statista~\cite{statista}, the overall income of the worldwide AI market is expected to reach 126 billion dollars by 2025.
Typically, AI services are data-hungry. The training of valuable AI models needs to collect and process large amounts of data. 
However, serious concerns are rising about the privacy and security of the collected training data.
Privacy regulations, such as EU's General Data Protection Regulation (GDPR), California Consumer Privacy Act (CCPA), and HIPAA, restrict the collection and direct use of data.
To this end, \textit{privacy-preserving distributed machine learning}~\cite{mcmahan2017communication, vepakomma18} has emerged to train AI models collaboratively at clients without revealing and transmitting the raw data.
The data and labels are kept locally at the clients to relieve the burden of privacy protection for the service providers.

Split learning is a promising privacy-preserving distributed machine learning framework, where the full AI model is cut on a layer-wise basis into multiple portions to be trained at the participants (e.g., clients and servers) collaboratively~\cite{gupta18, vepakomma18}.
the service provider (or server) can neither be informed of the raw data (only the intermediate results at the cut layer) nor have access to the client-side AI model.
As a result, split learning is regarded to be privacy-preserving due to its architectural design.
Recently, the effectiveness and efficiency of split learning have been studied via empirical experiments~\cite{vepakomma18, poirot19, koda19, singh19, abuadbba20, gao20, kim20, thapa21} and architecture designs~\cite{abedi20, turina20, ceballos20, jeon20, thapa20, palanisamy21, romanini21}.
As a promising privacy-preserving distributed learning paradigm, split learning has attracted attention from various privacy-sensitive applications, such as medical platforms for healthcare~\cite{vepakomma18, poirot19, ha21spatio}, camera surveillance~\cite{ha21secure}, and wireless communications~\cite{koda19, koda20communication}.

\begin{figure}[!t]
	\begin{center}
		\subfloat[Split learning with label sharing] {
			\begin{minipage}[t]{\linewidth}
				\includegraphics[width=0.9\linewidth]{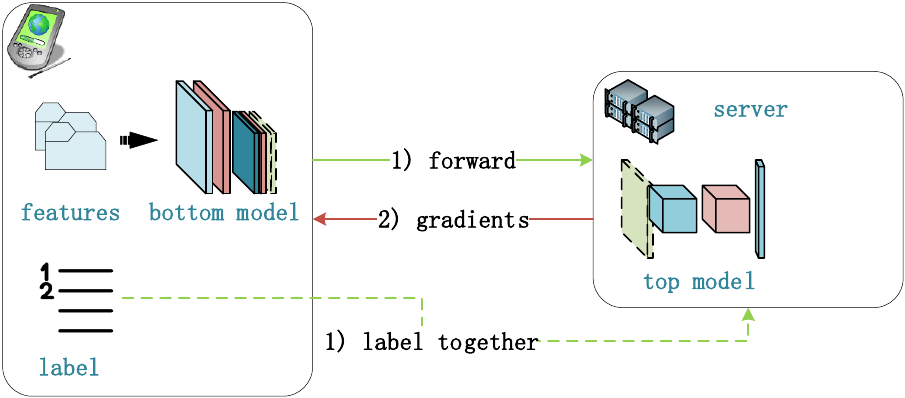}
				\label{fig:withlabel-split-learning}
			\end{minipage}
		}
		
		\subfloat[Vanilla split learning] {
			\begin{minipage}[t]{\linewidth}
				\includegraphics[width=0.9\linewidth]{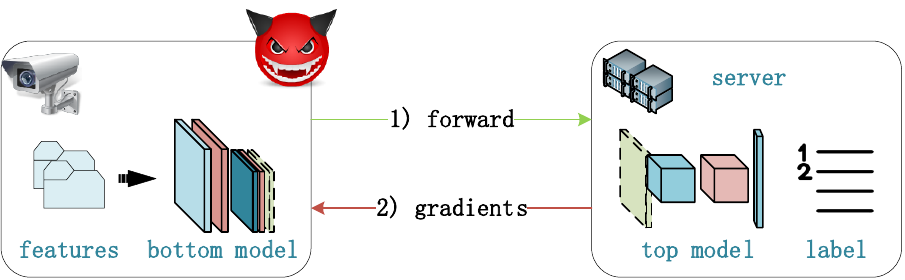}
				\label{fig:vallina-split-learning}
			\end{minipage}
		}
		
		\subfloat[Split learning without label sharing] {
			\begin{minipage}[t]{\linewidth}
				\includegraphics[width=0.9\linewidth]{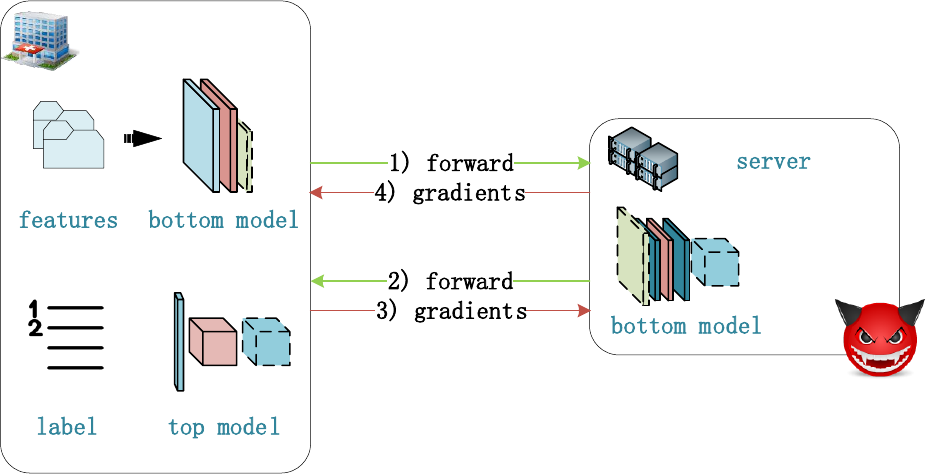}
				\label{fig:nolabel-split-learning}
			\end{minipage}
		}
		\caption{The illustration of basic split learning setups with and without label protections, where setups (b) and (c) are designed to protect the private labels.}
		\label{fig:split-learning-setup}
	\end{center}
	\vskip -0.2in
\end{figure}

Fig.~\ref{fig:split-learning-setup} summarizes the widely-adopted split learning setups with and without label protections. 
For ease of presentation, the learning model portions with/without the output layers are denoted by the top/bottom model in the rest of the paper. 
The basic steps of split learning are as follows.
1) \textit{Local Inference:} The client runs the forward-pass to the cut layer and reports the smashed data (i.e., features extracted from the raw data) to the server;
2) \textit{Server Inference and Training:} The server executes the inference on the remaining layers, computes the loss, trains the server-side model via \textit{back-propagation}, and sends the gradients of smashed data to the client;
and 3) \textit{Client Training:} The client finishes the training of the client-side model based on the received gradients.
The operations in Fig.~\ref{fig:split-learning-setup}\subref{fig:withlabel-split-learning} follow the standard steps above, where the client transmits the labels together with the smashed data to its trusted server.

Note that the trusted third-party service provider (or server) may not always be available in many practical applications. 
Directly sharing of private labels (e.g., patient medical records) would cause severe privacy and security concerns.
Moreover, the adversary can make free use of the recovered private labels and even reconstruct a complementary learning model.
To this end, split learning with label protections is shown in Figs.~\ref{fig:split-learning-setup}\subref{fig:vallina-split-learning} and \ref{fig:split-learning-setup}\subref{fig:nolabel-split-learning}.
In particular, in Fig.~\ref{fig:split-learning-setup}\subref{fig:vallina-split-learning}, private set intersection~(PSI) protocols~\cite{kolesnikov16, pinkas18} are adopted in vanilla split learning to align the data records before training, such that the labels can reside in the server.
In Fig.~\ref{fig:split-learning-setup}\subref{fig:nolabel-split-learning}, the learning model is divided into three portions, where the client has both the input and output portions of the model to locally evaluate the loss (to share gradients instead of the private labels).
The process requires additional client-server communication for label protections.

This paper aims to answer the following research questions, including 1) \textit{``Whether split learning with label protections can really protect the private labels,''} and 2) \textit{``Whether the attacks of split learning can be robust during both the training and inference phases and against different defense mechanisms.''}
Some recent studies~\cite{Yuan21, li22, erdogan21, kariyappa21} have reported that the adversary could recover the private labels against split learning under binary~\cite{Yuan21,li22} and multi-class~\cite{erdogan21,kariyappa21} classification tasks. 
However, it still lacks mathematical analysis of label leakages for multi-class classifications, which is critical for understanding the security performance of split learning.
Moreover, the existing multi-class label inference attacks~\cite{erdogan21,kariyappa21} may not be applicable under many practical settings (e.g., different positions of cut layers, lack of the victim's model architecture, and diverse training hyperparameters).

In this paper, we first mathematically analyze the possible label leakages and propose the similarity measurement for efficient label inference attacks in both label-protection cases. 
In the generalized threat model, the adversary can be either the server or client for Figs.~\ref{fig:split-learning-setup}\subref{fig:vallina-split-learning} and~\ref{fig:split-learning-setup}\subref{fig:nolabel-split-learning}, respectively.
We reveal that, even without the knowledge of learning models, the gradients or smashed data at the cut layer can already reveal private labels.
We propose the cosine and Euclidean similarity measurements for the gradients (attacks during the training phase) and smashed data (attacks during the inference phase), respectively.
Then, the two similarity measurements are shown to be unified in Euclidean space.
Based on the proposed similarity measurement, we design three inference attacks during both the training and inference phases of split learning. 
The key contributions are summarized as follows.
\begin{itemize}
	\item We mathematically analyze the vulnerabilities of split learning from three different types of gradients and smashed data. 
Both cosine and Euclidean similarity measurements are proposed and then unified in the Euclidean space to facilitate efficient label inference attacks.
	\item We propose three efficient similarity-based label inference attacks during both the training and inference phases of split learning.
The proposed attacks are easy-to-conduct and only require one auxiliary sample for each label class.
	\item We experimentally evaluate the attack performance under six different datasets. The proposed approach can achieve close to $100\%$ attack accuracy when the cut layer approaches the output.
	\item We also test the attack performance against various state-of-the-art defense mechanisms, including DP-SGD, label differential privacy, gradient compression, and Marvell.
	The proposed attacks are shown to be robust against the existing defenses\footnote{The
source code is available at the GitHub repository:\\
https://github.com/ZeroWalker10/sl\_similarity\_label\_inference.git.}.
\end{itemize}

In our previous work~\cite{liu23}, we have proposed a distance-based label inference attack for split learning.
This work is novel and different from the previous one in the following aspects: 1) Motivated by the proposed/analyzed similarity measurement, we propose the clustering-based label inference attack to exploit the grouping features of the collected data samples for better attack performance. 2) We validate the clustering feature of gradients and smashed data in a visualize manner and design the clustering attack algorithm by integrating K-means clustering and K-M matching algorithms~\cite{kuhn1955hungarian}. 3) We conduct more experiments by evaluating complicated dataset (e.g., ImageNet with 1000 different classes), robustness of the proposed attacks (different epochs, batch sizes, and labeled samples), new attack benchmarks and defense mechanisms.

The rest of this paper is organized as follows. 
Section II briefly summarizes the recent progress of inference attacks in distributed machine learning. 
In Section III, we present the proposed similarity measurements and three label inference attacks against both the training and inference phases of split learning.
Sections IV and V evaluate the performance of the proposed approach under different datasets and defense mechanisms, followed by the conclusion in Section VI.

\section{Related Work}
\label{sec:related}
This section provides a concise overview of the different distributed learning techniques, including distributed learning, federated learning, and split learning and discusses the current inference attacks that pose potential risks to privacy-preserving distributed learning.

\subsection{Preliminaries on Different Distributed Learning Techniques}
There are several different techniques to achieve efficient and privacy-preserving distributed learning across multiple network devices or organizations.
Three prominent techniques that have gained significant attention are distributed learning, federated learning, and split learning. 

1) Distributed learning enables efficient model training across multiple machines. 
Typically, a large volume of data is divided into subsets and distributed to the working nodes.
The same learning model is trained on the subsets parallelly, which is the data-parallel approach.
An alternative approach is model-parallel.
In particular, the copies of the entire dataset are applied to train different parts of a learning model by the working nodes~\cite{verbraeken2020survey}.

2) Federated learning is one privacy-preserving learning framework across multiple clients or organizations~\cite{mcmahan2016federated, verbraeken2020survey}. 
Instead of sending raw data, federated learning enables the training of models directly on the devices themselves. 
A parameter server is required to aggregate the local model updates from the devices to ensure global convergence.

3) Split learning is another promising privacy-preserving distributed machine learning framework, where the full AI model is cut on a layer-wise basis into multiple portions to be trained at the participants (e.g., clients and servers) collaboratively~\cite{gupta18, vepakomma18}. 
The procedure of split learning has been presented in Fig.~\ref{fig:split-learning-setup}.

\subsection{Summary of Inference Attacks against Privacy-preserving Distributed Learning}
As stated in Section~\ref{sec:intro}, split learning is still in its infancy stage with only limited work on its attack methods.
For comprehensive review and comparisons, we extend the topic to inference attacks in privacy-preserving distributed machine learning (including both federated learning and split learning).
Given the distinctive attack objectives, the existing work can be categorized into the inferences of 1) class representatives, 2) memberships and training data properties, and 3) training samples and labels, respectively.
The details are as follows.

\begin{table*}[ht]
	\caption{Comparison of Different Label Inference Attacks.}
	\label{tlb:methods-table}
	\vskip 0.15in
	\begin{center}
		\begin{scriptsize}
			\begin{sc}
				\begin{tabular}{lccccc}
					\toprule
					\multirow{5}{*}{Attack} & \multirow{5}{*}{\makecell{Multi-class\\Classification}} & \multicolumn{3}{c}{Characteristics of Split Learning} & \multirow{5}{*}{Highlights}\\
					\cmidrule(lr){3-5}
					& & \multicolumn{2}{c}{\makecell{Flexible Cut Layer}} &  \multirow{3}{*}{\makecell{Lack of\\Top Model}} & \\
					\cmidrule(lr){3-4}
					& & gradients & smashed & & \\
					\midrule
					Fu et al.~\cite{Fu22} & $\surd$ & $\times$  & $\surd$  & $\surd$ & \makecell{Training a surrogate top model to \\complete the model with tens of\\ labeled samples per class} \\
					\midrule
					Yuan et. al~\cite{Yuan21} & $\times$ & $\surd$  & $\times$ & $\surd$ & \makecell{Reconstructing the binary labels \\with linear equations}\\
					\midrule
					Li et al.~\cite{li22} & $\times$ & $\surd$ & $\times$ & $\surd$ & \makecell{Infering the binary labels with \\gradient norm or direction}\\
					\midrule
					Ege et al.~\cite{erdogan21} & $\surd$ & $\times$ & $\times$ & $\times$ & \makecell{Enumerating the labels with the top model \\to match the received gradients}\\
					\midrule
					Sanjay et al.~\cite{kariyappa21} & $\surd$ & $\times$ & $\times$ & $\surd$ & \makecell{Optimizing the labels through\\ replaying the collected gradients}\\
					\midrule
					Xie et al.~\cite{xie23} & $\surd$ & $\surd$ & $\times$ & $\times$ & \makecell{Optimizing the labels with a surrogate top model \\to match the received gradients}\\
					\midrule
					Our Work & $\surd$ & $\surd$ & $\surd$ & $\surd$ & \makecell{Recovering the labels during both training\\ and inference phases with a\\ unified similarity measurement and \\one labeled sample per class}\\
					\bottomrule
				\end{tabular}
			\end{sc}
		\end{scriptsize}
	\end{center}
	\vskip -0.1in
\end{table*}

\textbf{1) Inference of Class Representatives.}
The adversary synthesizes realistic-looking samples to extract private information (e.g., sensitive medical records) from honest participants by collecting and analyzing the model parameters.
Hitaj et al.~\cite{Hitaj17} proposed to apply Generative Adversarial Network~(GAN) to effectively launch the class representatives inference attack against distributed deep learning.
Wang et al.~\cite{wang19} attempted to further pursue the user-level representatives (beyond the typical global representatives).
The method in~\cite{wang19} combined GAN with a multi-task discriminator to achieve the simultaneous discrimination of category, reality, and client identity based on the input samples.

\textbf{2) Inference of Membership and Properties.}
The adversary aims to determine the membership of a random data sample (i.e., whether the data sample is in the model's training dataset) or infer the data sample properties maintained at the other participants.
For membership inference, Shokri et al.~\cite{shokri17} designed to train an attack model to classify the data sample memberships based on the black-box query results. Truex et al.~\cite{truex19} developed a general framework for membership inference attacks, and further investigated the model vulnerability against the attacks.
For property inference, Shen et al.~\cite{Meng21} proposed an active property inference attack method to retrieve the property from the model updates in blockchain-assisted federated learning.
Dynamic participant selection was also designed to further improve the attack accuracy in a large number of devices~\cite{Meng21}.

The joint inference of membership and properties was studied in \cite{melis19}, where the non-zero gradients of the embedding layer were shown to reveal the positions of the words within a training batch.
As a result, the adversary can perform membership inference by exploiting the embedding layer gradients.
Melis et al.~\cite{melis19} also developed a passive property inference attack.
The adversary can compute the aggregated model updates based on the snapshots of the global model and train a binary property classifier to infer the targeted property.

\textbf{3) Inference of Training Samples and Labels.}
The adversary reconstructs the original training samples or the data labels from the intermediate information~\cite{lyu23} (e.g., model gradients and smashed data).
Label inference in split learning is the focus of this paper.
In the following, we summarize the existing work on the topics of federated learning and split learning, respectively.

In the process of federated learning, the adversary can have access to the gradients of other devices.
The existing work applied either optimization techniques (e.g., gradient matching~\cite{zhu19}) or mathematical analysis (e.g., gradient properties~\cite{zhao20,wainakh21} and singular value decomposition~(SVD)~\cite{dang21}) to infer the private labels.
In particular, Zhu et al.~\cite{zhu19} proposed to match the received gradients to achieve pixel-wise accurate data recovery for images.
The authors in~\cite{zhao20,wainakh21,dang21} analyzed the gradients of the last hidden layer to determine the data labels based on the signs~\cite{zhao20}, the directions and magnitudes~\cite{wainakh21}, and the SVD value~\cite{dang21}.

As a promising and emerging technique for distributed learning, split learning has attracted recent work on the inference attacks of training samples~\cite{pasquini21} and data labels~\cite{erdogan21,Yuan21,li22,kariyappa21}.
Dario et al.~\cite{pasquini21} showed that a malicious server could hijack the learning process of split learning to infer clients' training samples.
The authors in~\cite{Yuan21,li22} proposed the label inference approaches for binary classification.
In particular, Yuan et al.~\cite{Yuan21} designed deterministic attacks for private labels by solving linear equations.
Li et al.~\cite{li22} analyzed the gradient properties for label attacks of binary classification.

The most relevant studies are label inferences for multi-classification tasks in split learning~\cite{erdogan21,kariyappa21,Fu22,xie23}.
Ege et al.~\cite{erdogan21} considered that the adversary knew the victim's model structure and could reconstruct an artificial top model to determine the data label by matching the received gradients and the gradients of the artificial top model.
Sanjay et al.~\cite{kariyappa21} designed a gradient inversion attack by formulating the label leakage attack as a supervised learning problem with the input of smashed data and gradients.
However, the adversary in~\cite{kariyappa21} must collect all the gradients and smashed data and have the full knowledge of the private label distributions, which may not be available in practice.
Fu et al.~\cite{Fu22} proposed two label inference attack methods by analyzing the signs of gradients at the output layer or {constructing a complementary top model in a semi-supervised manner.
Xie et al.~\cite{xie23} optimized the private labels based on a surrogate top model to match the received gradients.

We summarize and compare the existing work on label inference attacks for split learning in Table~\ref{tlb:methods-table}.
Distinctively different from the state-of-the-art, this paper proposes label inference attacks for split learning from both gradients (\textit{during training}) and smashed data (\textit{during inference}).
Without the knowledge of the victim's top model, the proposed attacks are robust under different settings and defense mechanisms.

\section{Label Inference in Split Learning}
\label{sec:label-leakage}
\begin{figure*}[ht]
	\begin{center}
		\centerline{\includegraphics[width=\textwidth]{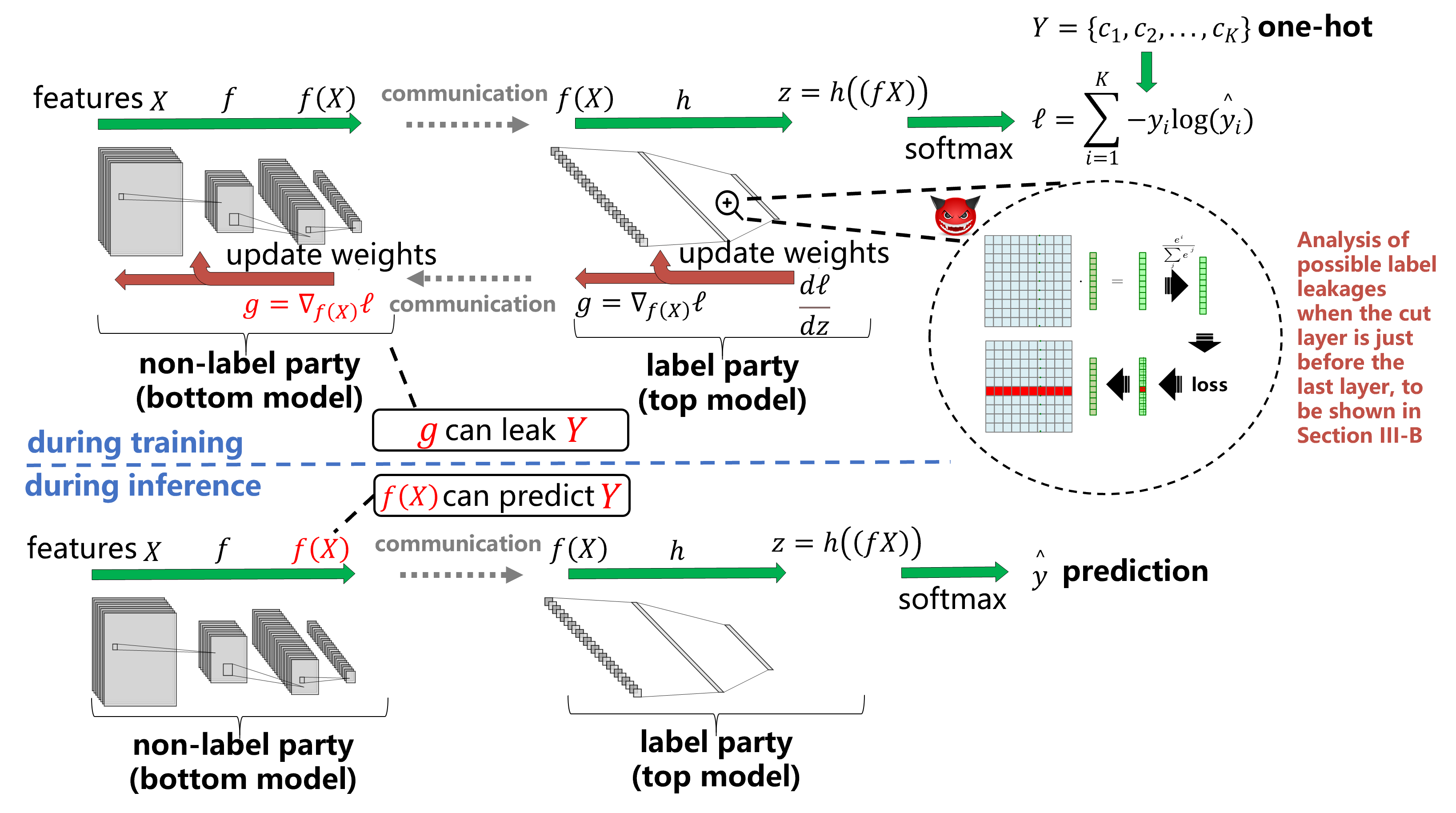}}
		\caption{Threat model of split learning. During training, gradient $g$ can leak private label information. During inference, smashed data $f(X)$ can also disclose private labels}.
		\label{fig:scenes}
	\end{center}
	\vskip -0.2in
\end{figure*}

This paper considers label-protection split learning scenarios in Figs.~\ref{fig:split-learning-setup}\subref{fig:vallina-split-learning} and~\ref{fig:split-learning-setup}\subref{fig:nolabel-split-learning}.
Without loss of generality, we summarize the threat model of both scenarios in Fig.~\ref{fig:scenes}. The malicious/non-label party (either the client in Fig.~\ref{fig:split-learning-setup}\subref{fig:vallina-split-learning}  or the server in Fig.~\ref{fig:split-learning-setup}\subref{fig:nolabel-split-learning}) can have access to the smashed data and gradients to perform the label inference attack.
In the following, we illustrate the threat model in Section~\ref{subsec:threat}, mathematically analyze the potential label leakages and propose the cosine and Euclidean similarity measurements in Section~\ref{subsec:possible-leakage}, and demonstrate the proposed label inference attack approach in Section~\ref{subsec:attack}.

\subsection{Threat Model}\label{subsec:threat}
We consider the honest-but-curious attack model for the malicious/non-label party (also denoted as the adversary in the rest of the paper) in split learning.
The adversary follows the standard steps of split learning (as stated in Figs.~\ref{fig:split-learning-setup}\subref{fig:vallina-split-learning} and~\ref{fig:split-learning-setup}\subref{fig:nolabel-split-learning}), and performs the passive label inference attacks based on the smashed data and gradients without interfering with the split learning procedure.
As a curious-but-honest adversary, the non-label party naturally participates in the split learning process, and can have access to the cut-layer information (gradients and smashed data) for attacks. 
Such threat model is widely adopted in the literature. 
For example, the non-label party can use the magnitudes and directions of the cut-layer gradients to infer the private labels~\cite{li22}.

In particular, the adversary conducts the label inference attack with the following steps:
1) the adversary reports the smashed data (i.e., the features extracted from the raw data via running the forward pass of the local bottom model) to the victim;
2) the victim calculates the loss function based on the factual labels, updates its local (i.e., top) model via \textit{back-propagation}, and sends the gradients to the adversary; and
3) the adversary updates the bottom model based on the received gradients.

Note that the adversary does not modify the operations of split learning.
The victim can hardly detect the passive label inference attack and may only choose to adopt the existing privacy defense techniques for protection.
As will be shown in the experimental results, the proposed attack approach is robust against the typical defense methods.
Fig.~\ref{fig:scenes} plots the standard training and inference process in a two-party split learning.
The label inference attacks can be conducted both during and after the training
process of split learning.
During training, the non-label party can exploit the cut-layer gradients from the label party to infer private labels, as the received gradients already carry information about private labels.
After training, the trained submodel at the non-label party has learned good representation of the inputs with the help of private labels.
Thus, the extracted embeddings also contain information about private labels and can be exploited to infer private labels.

We summarize the application settings and the corresponding input data as follows.
\begin{enumerate}
	\item \textbf{Gradient Attack during Training Phase.}
	During the back-propagation process, the adversary can collect the gradients of smashed data received from the victim (i.e., the label party).
	The proposed attack method can recover the private label from the collected gradients, as shown at the top of Fig.~\ref{fig:scenes}.
	\item \textbf{Smashed Data Attack during Inference Phase.}
	In the absence of gradients after the training phase, the adversary itself can generate and collect the smashed data (features extracted via running the forward-pass of the bottom model) given random data samples for label inference attack, as shown at the bottom of Fig.~\ref{fig:scenes}.
\end{enumerate}

The label inference attack is designed against the general multi-classification applications, aiming to minimize the cross-entropy loss function of the classification results.
The adversary has the a-priori knowledge of the label list of the classification applications and one labeled sample for each class.
The notations used in the rest of this paper are summarized as follows.
\begin{itemize}
	\item \textbf{\emph{$X$}} denotes the input features of the bottom model.
	\item \textbf{\emph{$y$}} and \textbf{\emph{$\hat{y}$}} are the ground-truth one-hot label and the corresponding prediction probability distribution of the learning model, respectively.
	\item \textbf{\emph{$c$}} is the ground truth label.
	\item \textbf{\emph{$W$}} is the weights of all the neurons, and $W_L$ denotes the weights of the neurons at the $L$-th layer of the model.
	\item \textbf{\emph{$a_L$}} is the activation output at the $L$-th layer of the model.
	\item \textbf{\emph{$z$}} is the output vector at the last hidden layer, also referred to as $logits$. The subscript is the index of $logits$. For instance, $z_i$ is the $i$-th item of $z$.
	\item \textbf{\emph{$\ell$}} is the loss of the model. The cross-entropy loss is defined as $\ell = -\sum\nolimits_{i}{y_ilog\hat{y}_{i}}$, where $\hat{y}_{i} = \frac{e^{z_i}}{\sum\nolimits_{j}e^{z_j}}$.
	\item \textbf{\emph{$\nabla$}} is the notation for gradients. Let $\nabla W$, $\nabla a$ and $\nabla z$ denote the gradients of the weights, activations and logits, respectively.
\end{itemize}

\subsection{Analysis of Possible Label Leakages}
\label{subsec:possible-leakage}

\begin{figure}[t]
	\vskip 0.2in
	\begin{center}
		\centerline{\includegraphics[width=\columnwidth]{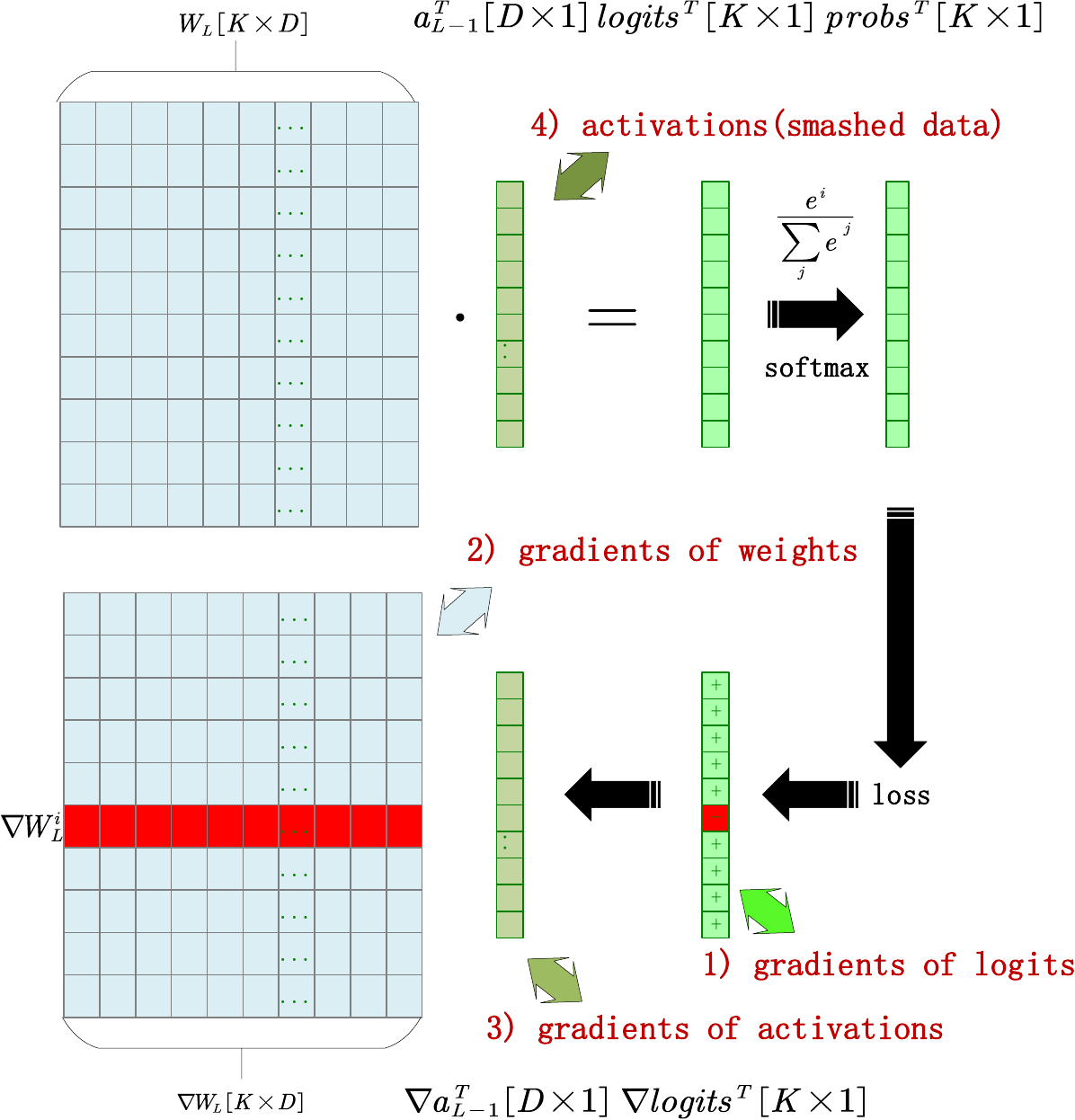}}
		\caption{Possible label leakages in split learning with an emphasis on the last hidden layer. $W_L$ is the weights of the last hidden layer and $a_{L-1}$ is the activation output of layer $L-1$. $\nabla W_L^i$ is the gradients of weights $W_L^i$ in terms of the $i$-th logit. $\nabla W_L$, $\nabla a_{L-1}$, and $\nabla logits$ denote the gradients of $W_L$, $a_{L-1}$ and $logits$ respectively.}
		\label{fig:possible-label-leakage}
	\end{center}
	\vskip -0.2in
\end{figure}

This section analyzes the possible label leakages (i.e., the attack means) in split learning.
We start by illustrating how the gradients of the last layer may reveal the private labels (as also done in the existing work~\cite{Fu22}).
Then, we extend the settings to the gradients/smashed data at previous layers to accommodate the general cases in practical split learning. We propose the cosine and Euclidean similarity measurements for our attack methods.

Fig.~\ref{fig:possible-label-leakage} depicts the \textit{forward-pass} and \textit{back-propagation} in the last hidden layer for a typical classification task.
In particular, the activations/smashed data $a_{L-1}$ are extracted via the forward-pass of the learning model.
The label party can calculate the cross-entropy loss function after receiving $a_{L-1}$.
Given the loss value, during the \textit{back-propagation}, three different types of gradients are generated with respect to the logits $z$, weights $W$, and activations $a_{L-1}$, respectively.
In summary, there are four possible ways of label leakages(as shown in Fig.~\ref{fig:possible-label-leakage}), including
1)  the gradients of logits, 2) the gradients of weights, 3) the gradients of activations, and 4) smashed data.

In the following, we first briefly introduce the leakage analysis of the first two types of data, i.e., 1) the gradients of logits and 2) the gradients of weights.
The first two types of data are used for inference attacks in~\cite{Fu22,zhao20} which may not be applicable to practical split learning.
Then, we propose the cosine and Euclidean similarity measurements for 3) the gradients of activations and 4) smashed data.
The gradients of activations and smashed data are shown to have grouping features (based on the proposed measurements) to be shown in Fig.~\ref{fig:samples-clustering}. This serves as the motivation of our proposed attacks against split learning.
The details are as follows.

\textbf{1) The Gradients of Logits.}
The partial derivative of $z_i$ with respect to the loss function $\ell$ can be given by
\begin{equation}
	\label{eq:logit-gradient}
	\frac{\partial \ell}{\partial z_i} = \begin{cases}
		\hat{y}_i - 1, &i = c\text{;}\\
		\hat{y}_i - 0, &i \neq c\text{.}
	\end{cases}
\end{equation}
As a result, the gradients of the logits $\nabla z$ can be denoted as $\frac{\mathrm{d} \ell}{\mathrm{d} z} = \hat{y} - y$.
We can see that $\nabla z_i \in (-1, 0)$ when $i = c$ and $\nabla z_i \in (0, 1)$ when $i \neq c$ (given that $\hat{y}_i = \frac{e^{z_i}}{\sum_j e^{z_j}} \in (0, 1)$).
To this end, the ground-truth label can be determined by finding the index of the negative element in $\nabla z$.
Fu et al.~\cite{Fu22} launched the label inference attack based on the observation.
However, the attack method requires the accessibility of the final layer in a learning model, which may not be available in practical split learning.

\textbf{2) The Gradients of Weights.} Zhao et al.~\cite{zhao20} relaxed the assumption of knowing the gradients of $logits$, and considered that the adversary could only have the gradients of the weights in the last hidden layer.
Let $\nabla W_L^i$ be the gradient with respect to the weights $W_L^i$, which can be given by
\begin{equation}
	\begin{split}
		\nabla W_L^{i} = \frac{\partial \ell}{\partial W_L^{i}} &= \frac{\partial \ell}{\partial z_i} \cdot \frac{\partial z_i}{\partial W_L^i} \\
		&= \nabla z_i \cdot \frac{\partial (a_{L-1}{W_L^i}^T + b_L^i)}{\partial W_L^i} \\
		&= \nabla z_i \cdot a_{L-1}\text{;}
	\end{split}
    \label{eq:weg_grad}
\end{equation}
where $a_{L-1}$ is the activation output of layer $L-1$, and $b_L^i$ is the bias parameter.
The activation vector $a_{L-1}$ is independent of the logit index $i$. According to Eq.~\ref{eq:weg_grad} the ground-truth label $c$ can be identified based on the sign of $\nabla W_L^i$, as given by
\begin{equation}
	c = i, \text{if } \nabla W_L^i \cdot \nabla {W_L^j}^T \leq 0, \forall j \neq i\text{.}
\end{equation}
However, the gradients of weights still require access to the last layer, and may not be suitable for practical split learning.

\begin{figure*}[ht]
	\vskip 0.2in
	\begin{center}
		\subfloat[Gradients of last layer] {
			\begin{minipage}[t]{0.23\linewidth}
				\includegraphics[width=0.9\linewidth]{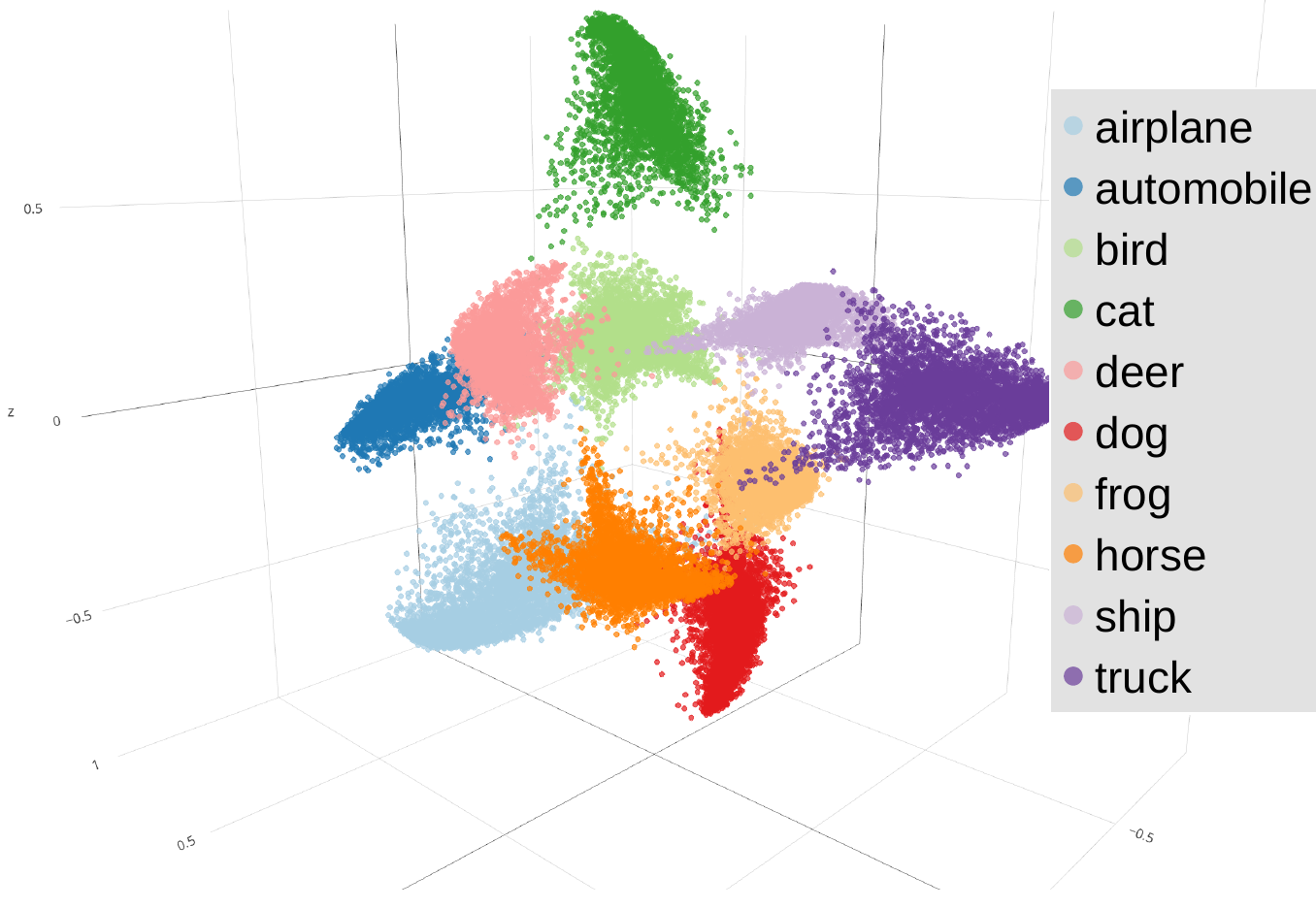}
				\label{fig:cifar10-last-layer}
			\end{minipage}
		}
		\subfloat[Smashed Data of last layer] {
			\begin{minipage}[t]{0.23\linewidth}
				\includegraphics[width=0.9\linewidth]{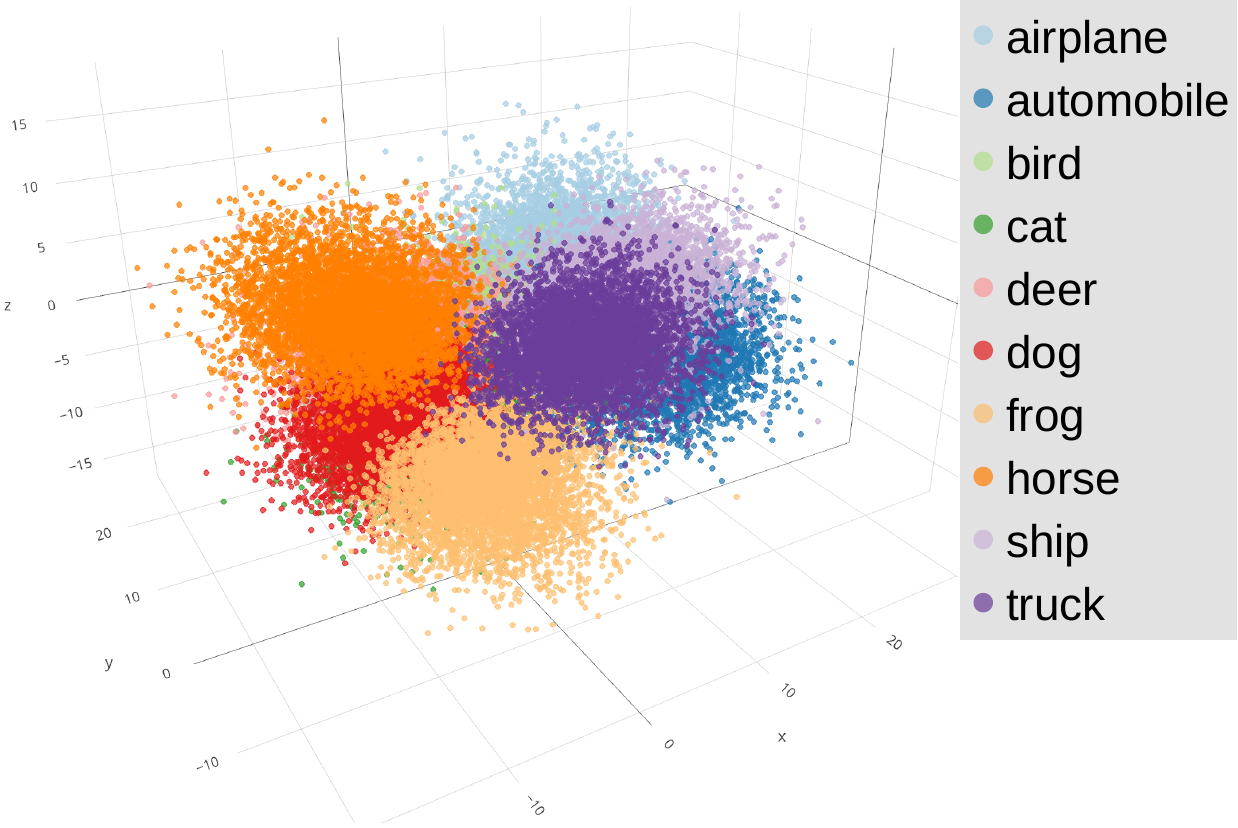}
				\label{fig:cifar10-last-layer-smashed}
			\end{minipage}
		}
		\subfloat[Gradients of reciprocal third layer] {
			\begin{minipage}[t]{0.23\linewidth}
				\includegraphics[width=0.9\linewidth]{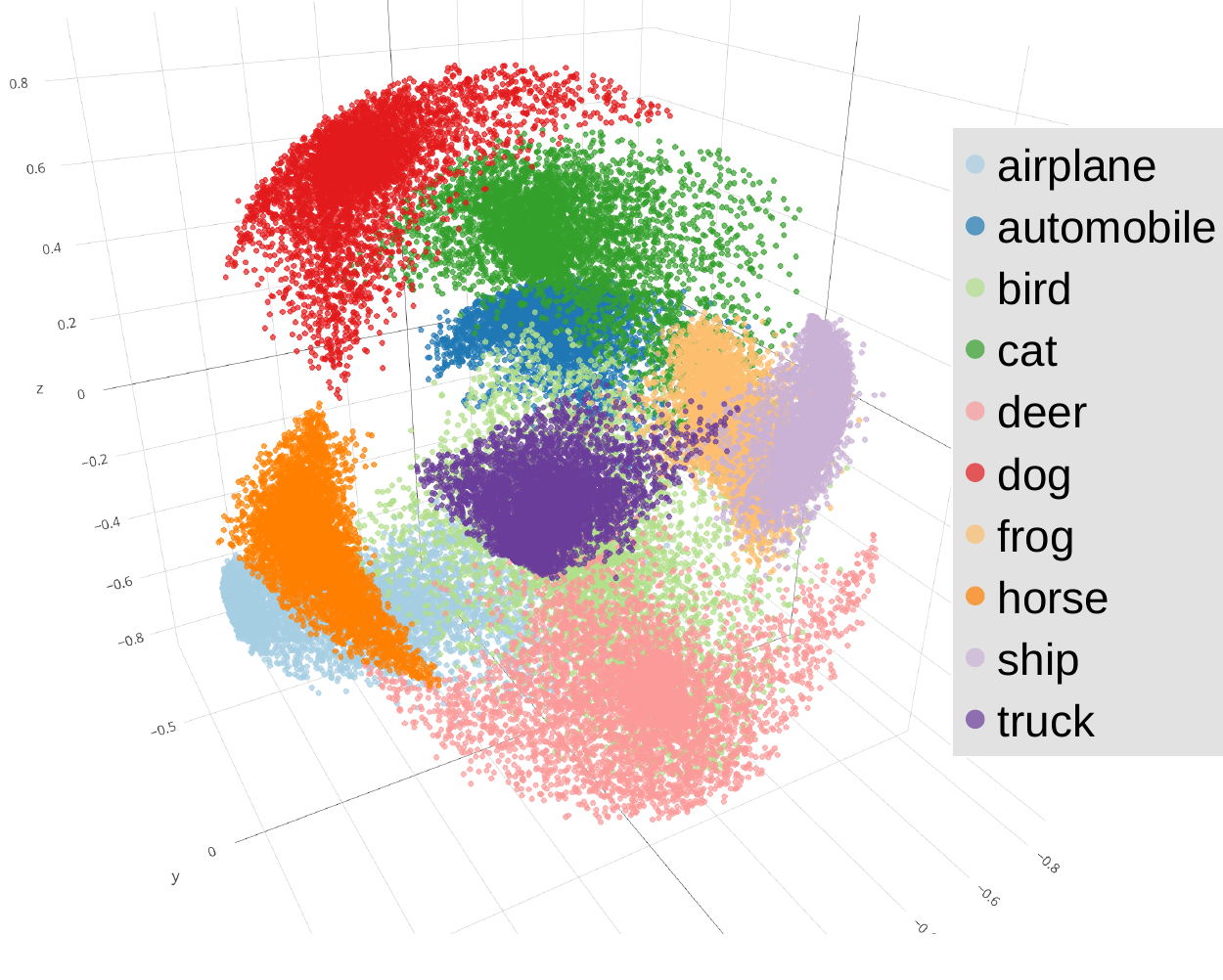}
				\label{fig:cifar10-3rd-layer}
			\end{minipage}
		}
		\subfloat[Smashed Data of penultimate layer] {
			\begin{minipage}[t]{0.23\linewidth}
				\includegraphics[width=0.9\linewidth]{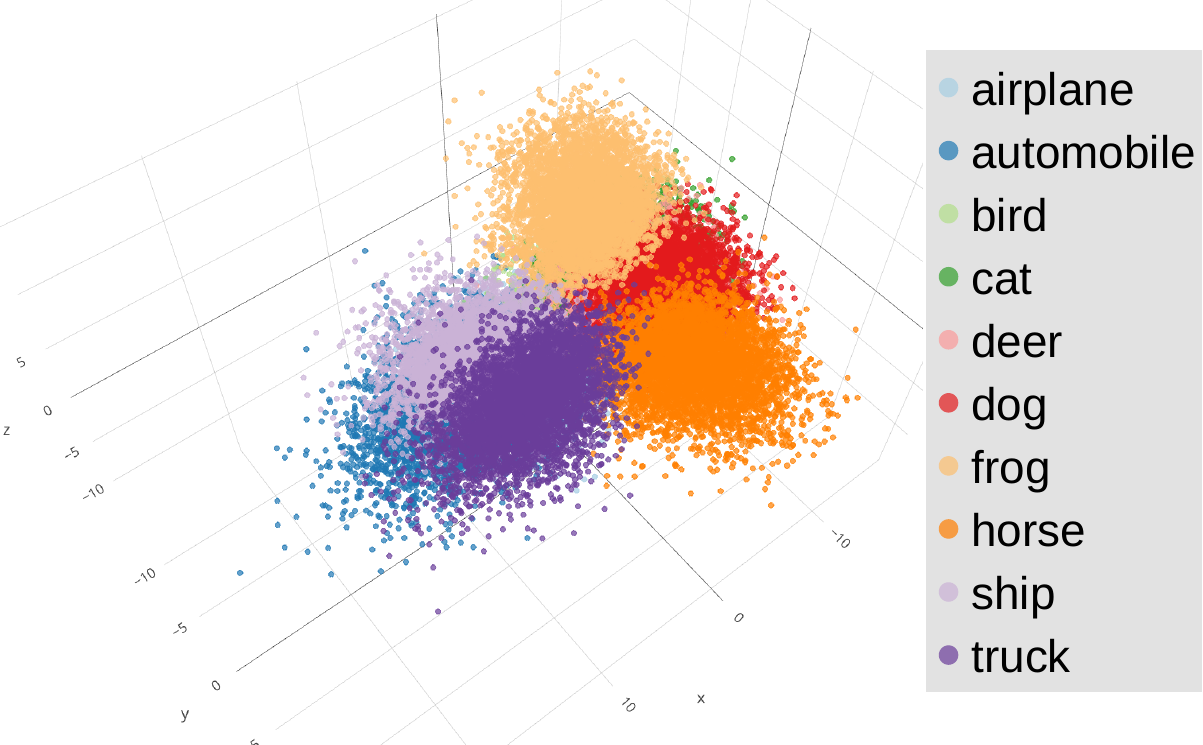}
				\label{fig:cifar10-2nd-layer-smashed}
			\end{minipage}
		}
		\caption{Samples distributions on CIFAR-10. The captions of subfigures demonstrate the dataset and the position of the cut layer.}\label{fig:samples-clustering}
	\end{center}
	\vskip -0.2in
\end{figure*}

\textbf{3) The Gradients of Activations.}
Consider the practical setting of split learning, where only the gradients of the activations at the cut layer are available.
The $i$-th element of the gradients of the activations of layer $L-1$, denoted by $\nabla a_{L-1}^i$, can be given by
\begin{equation}
	\begin{split}
		\nabla a_{L-1}^i &= \frac{\partial \ell}{\partial a_{L-1}^i} = \frac{\mathrm{d} \ell}{\mathrm{d} z} \frac{\partial z}{\partial a_{L-1}^i}\\
		&= (\hat{y} - y) \frac{\partial (a_{L-1} {W_L}^T + b_L)}{\partial a_{L-1}^i}
= (\hat{y} - y) {{W_L^i}}^T \text{.}
	\end{split}
	\label{eq:gradient-last-layer}
\end{equation}
As a result, the gradients of the activations at layer $L - 1$ can be written as $\nabla a_{L-1} = (\hat{y} - y) {W_L}^T$. \footnote{We also highlight that the gradients of activations can be easily extended to other layers for split learning.
	Let $\nabla a_{i}$ be the gradients of activations at the $i$-th layer from the output layer, which can be written as $\nabla a_{i} = \frac{\mathrm{d} \ell}{\mathrm{d} z} \frac{\mathrm{d} z}{\mathrm{d} a_i}=(\hat{y} - y) \frac{\mathrm{d} z}{\mathrm{d} a_i}$. $\nabla a_{i}$ exhibits the same structure as $\nabla a_{L-1}$ in Eq. \eqref{eq:gradient-last-layer}. For brevity, we design the similarity based on $\nabla a_{L-1}$.}

Note that the gradients of the activations at layer $L-1$ in Eq. \eqref{eq:gradient-last-layer} cannot directly reveal the private labels by comparing the signs/values of the gradients (as done in the gradients of the first two cases).
We proceed to show that the gradients from the data samples of the same class intend to resemble each other, and propose the cosine similarity metric to measure the resemblance.

\textit{Cosine Similarity Measurements}.
Let $\nabla a_{L-1}$ and $\nabla a_{L-1}^{\prime}$ be the gradients of activations at layer $L-1$ from any two data samples of the classes $i$ and $j$, respectively.
$y$/$\hat{y}$  and $y^\prime$/$\hat{y}^\prime$ are the ground-truth one-hot label and predicted probability distribution for the samples from classes $i$ and $j$, respectively.
$cos(i,j)$ denotes the cosine similarity measurement of the two data samples.
Given the same neuron weights ${W_L}^T$ within a training batch\footnote{The cosine similarity is also applicable between different batches(as also shown in Fig.~\ref{fig:samples-clustering} and the experimental results). Let $\nabla a_{L-1}^{\prime}$ be the gradients of activations at the next training batch. $\vert B \vert$ and $\eta$ denote the batch size and learning stepsize, respectively.
	We have $\nabla a_{L-1}^{\prime} = (\hat{y} - y) {W_L^{\prime}}^T = (\hat{y} - y)(W_L - \frac{\eta}{\vert B \vert} \sum \nabla W)=(\hat{y} - y)W_L - (\hat{y} - y)\frac{\eta}{\vert B \vert} \sum \nabla W$.
	The second term can negligible with typically small stepsize $\eta$.}, the resemblance of $\nabla a_{L-1}$ and $\nabla a_{L-1}^{\prime}$ can be measured by the similarity between $\hat{y}-y$ and $\hat{y}^{\prime}-y^{\prime}$.
As a result, the cosine similarity $cos(i,j)$ is equivalent to $cos((\hat{y}-y), (\hat{y}^{\prime}-y^{\prime}))$, and can be given by
\begin{equation}\label{eq:cosine}
	\begin{aligned}
		cos(i,j) &= \frac{(\hat{y}-y) \cdot (\hat{y}^{\prime}-y^{\prime})}{\Vert \hat{y}-y \Vert \cdot \Vert \hat{y}^{\prime}-y^{\prime} \Vert}\\
		&= \alpha (\hat{y}-y) \cdot (\hat{y}^{\prime}-y^{\prime})\\
		&= \alpha (\hat{y} \cdot \hat{y}^{\prime} - \hat{y} \cdot y^{\prime} - y \cdot \hat{y}^{\prime} + y \cdot y^{\prime})\\
		&= \begin{cases}
			\alpha (\hat{y} \cdot \hat{y}^{\prime} - \hat{y} \cdot y^{\prime} - y \cdot \hat{y}^{\prime}) + \alpha, &i = j\text{;}\\
			\alpha (\hat{y} \cdot \hat{y}^{\prime} - \hat{y} \cdot y^{\prime} - y \cdot \hat{y}^{\prime}) + 0, &i \neq j\text{.}
		\end{cases}
	\end{aligned}
\end{equation}
where $\alpha = \frac{1}{\Vert \hat{y}-y \Vert \Vert \hat{y}^{\prime}-y^{\prime} \Vert}$ for brevity. 
We can see from Eq. \eqref{eq:cosine} that the gradients of the same class (i.e., in the case of $i=j$) would exhibit higher cosine similarity measurements than that of the different classes when $i \neq j$.
As a result, the cosine similarity measurements provide an efficient metric to extract the private labels by evaluating the similarities of the gradients of activations.

\textbf{4) Smashed Data.}
The values of gradients would diminish with the training procedure (e.g., the gradients may be increasingly close to zero when the learning model reaches convergence).
To this end, we continue to analyze the features of the smashed data (extracted from the raw data via forward pass) for label leakage after finishing the training procedure.
The classification-oriented learning model would increasingly extract more discriminative features of the input data with the increasing depth of the layers~\cite{Zeiler14}.
Then, the extracted features (or smashed data) can be used for the prediction of classification via the output layer.
A well-trained learning model can achieve superb extraction of input features and generate accurate predictions of the samples.

The extracted features (or smashed data) of a well-trained learning model have been shown and validated in \cite{Zeiler14}.
The adversary can also infer the private label by exploiting the features of smashed data.
For example, the adversary fine-tuned/trained a complementary bottom model to conduct the label inference attack based on the smashed data and auxiliary labeled data samples~\cite{Fu22}.
It should also be noted that the accuracy of the label inference attack from smashed data(features extracted from raw data) depends on the quality of the learning model (i.e., the performance of feature extraction).
In other words, the inference from smashed data may not be suitable during the training phase when the learning model is far from convergence.

\textit{Euclidean Similarity Measurements}. In this paper, we propose the label inference attack based on the similarity of the extracted features.
This is based on the simple observation that the samples of the same class may be more likely to have similar extracted features.
We adopt the Euclidean distance to measure the similarity of smashed data. As will be shown in the experimental results in Section~\ref{sec:experiment}, the attack is increasingly efficient when the cut layer approaches the output layer of the learning model.

\textbf{5) Visual Validation.}
Based on the cosine and Euclidean similarity measurements, the gradients of activations and smashed data are clustered according to their labels. The grouping features can be exploited for efficient label inference attacks in split learning.
To validate the analysis results, we collect the gradients of activations and smashed data from  CIFAR-10~\cite{Krizhevsky09} and visualize the results in Fig.~\ref{fig:samples-clustering}.

We normalize the collected gradients, reduce the vector of each data sample into three dimensions via principal component analysis~(PCA), and plot the 3D diagrams in Fig.~\ref{fig:samples-clustering}.
The normalization is not executed for plotting the smashed data.
Fig.~\ref{fig:samples-clustering} validates that both the gradients of activations and the smashed data exhibit the grouping features.

\subsection{Label Inference Attack}
\label{subsec:attack}
\begin{algorithm}[tb]
	\caption{The Proposed Clustering Label Inference Attack for Split Learning}
	\label{alg:cluster-attack}
	\begin{algorithmic}
		\STATE {\bfseries Input:} collected dataset $D$, gradients or smashed data of labeled samples $C$  /* $C \subset D$ */
		
		\STATE /* Initialzation */
		\STATE Normalize $D$
		\STATE Initialize centroids $U=\{\mu_1, \cdots\}$ with $C$
		
		\STATE
		\STATE /* $K\textnormal{--}means$ Algorithm */
		\REPEAT
		\FOR{every $d_i$ of $D$}
		\STATE $c_i = \mathop{\arg\min}\nolimits_{j}{\Vert d_i - \mu_j \Vert_2^2}$.
		\ENDFOR
		
		\FOR{each $\mu_j$ of $U$}
		\STATE $\mu_j = \frac{\sum_{i}{1\{c_i=j\}d_i}}{\sum_{i}{1\{c_i=j\}}}$.
		\ENDFOR
		\UNTIL{convergence}
		
		\STATE
		\STATE /* map cluster labels and ground truth labels */
		\STATE Extract the corresponding cluster labels of samples in $C$ and construct a set $\mathcal{C}$
		\STATE Map $C$ and $\mathcal{C}$ with K-M algorithm~\cite{kuhn1955hungarian}
		
		\STATE
		\STATE /* Reconstruct private labels */
		\FOR{every sample $a$ in $\mathcal{C}$}
		\FOR{every sample $b$ in $D$}
		\IF{cluster label of $a$ = cluster label of $b$}
		\STATE Assign corresponding ground-truth label of $a$ to the sample $b$
		\ENDIF
		\ENDFOR
		\ENDFOR
	\end{algorithmic}
\end{algorithm}

This section demonstrates the proposed label inference attacks for split learning.
Note that the similarities of the gradients of activations and smashed data are measured by the cosine similarity and Euclidean distance, respectively.
For the efficiency of label inference attacks, we show that these two metrics are interchangeable under normalized data samples to develop a unified attack method for both the gradients and smashed data.
In particular, the Euclidean distance of normalized vectors $x$ and $y$ (with $\vert x \vert = \vert y \vert = 1$) is
\begin{equation}
	\begin{split}
		\Vert x - y \Vert^2 &= \Vert x \Vert^2 + \Vert y \Vert^2 - 2x \cdot y\\
		&= 2 - 2cos(x, y)\text{.}
	\end{split}
	\label{eq:euclidean-distance}
\end{equation}
We can see from Eq. \eqref{eq:euclidean-distance} that maximizing the cosine similarity $cos(x, y)$ is equivalent to minimizing the Euclidean distance $\Vert x - y \Vert$.
As a result, in both the cases of the gradients of activations and smashed data, the similarity measurements can be unified as the Euclidean distance, where the data points are preprocessed to be normalized in the case of gradients.

Based on the unified Euclidean-distance similarity measurement, we propose three different label inference attack approaches based on the capability and available size of data samples at the adversary.
\textit{1) Euclidean-distance-based label inference:} The adversary is computationally resource-constrained and infers the private labels directly by finding the nearest Euclidean distance.
\textit{2) Clustering label inference:} The adversary can collect the data samples during the training and inference phase. Then, the adversary clusters the data samples to exploit the volume diversity for better attack performance.
\textit{3) Label inference with transfer learning:} With the input of original data, the adversary can adopt the transfer learning to extract the features and then conduct the similarity-based label inference attack.
The details of the three attack approaches are as follows.

\textit{1) Euclidean-distance-based Label Inference.} The adversary has some auxiliary data, i.e., a set of the
smashed data or gradients with labeled samples, denoted by $C$. The size of the labeled sample point set is the number of classes in the classification task (with one sample per class).
After receiving the gradient/smashed data $d_i$
, the adversary can assign it to its nearest sample point $c_j$, where $j=\underset{k}{\arg\min}\Vert d_i - c_k \Vert^2$ and $d_i$ is most likely to have the same label as $c_j$. 

\textit{2) Clustering Label Inference.} The proposed clustering label inference attacks are designed to group the data points (i.e., the collected gradients or smashed data) into $K$ clusters with similar characteristics, where $K$ is the number of classes of the classification task.
The clustering process is to determine $K$ centroids (i.e., the representatives of $K$ clusters) and assign the data points to the centroids (or clusters).
The objective is to minimize the within-cluster sum of similarity measurements (i.e., based on the Euclidean distance).
The assignment of the data points also reveals the private labels of the original data.

The clustering process can be categorized as the well-known $K\textnormal{--}means$ clustering algorithm in unsupervised learning.
Let $G = \{d_1, d_2, \cdots, d_m\}$ and $U = \{\mu_1, \mu_2, \cdots, \mu_k\}$ denote the collected data points and $K$ centroids of the clustering process, respectively.
The objective can be written as
\begin{equation}\label{eq:K-means}
	\min \quad E = \sum\nolimits_{i=1}^k{\sum\nolimits_{d \in c_i}{\Vert d - \mu_i \Vert_2^2}} \text{,}
\end{equation}
where $\mu_i = \frac{1}{ \vert c_i \vert}\sum_{d \in c_i}d$ is the centroid of cluster $i$, and $c_i$ is the assignment results containing all the data points within cluster $i$.
The $K\textnormal{--}means$ algorithm operates in an iterative manner, i.e., 1) assigning data points to minimize the within-cluster distance in Eq. \eqref{eq:K-means}, and 2) updating the centroids based on the assigned data points of each cluster.

We summarize the proposed clustering label inference attack for split learning in \textbf{Algorithm~\ref{alg:cluster-attack}}.
1) In the input phase, the adversary collects the gradients/smashed data, normalizes the gradients, and initializes the centroids with the auxiliary data (only one labeled data sample of each class).
2) In the clustering phase, the algorithm follows the standard $K\textnormal{--}means$ steps.
3) In the output phase, the adversary has the actual labels of labeled samples and their corresponding cluster labels. The attacks can map these two sets with the K-M algorithm~\cite{kuhn1955hungarian}. Finally, the cluster labels of the collected data can be assigned the mapped actual labels.

\textit{3) Label Inference with Transfer Learning.} As stated above, the quality of feature extraction (of the smashed data) is critical to the accuracy of label inference attacks. 
As a result, the adversary can leverage the published models of similar tasks to extract the features. 
Various model repositories, e.g., Model Zoo\footnote{\url{https://modelzoo.co/}} and PyTorch Hub\footnote{\url{https://pytorch.org/hub/}}, offer a collection of pre-trained models for utilization.
This enables potential adversaries to conveniently access and download pre-trained models relevant to similar tasks, facilitating the execution of transfer learning-based inferences.
After obtaining the pre-trained model, the adversary can exploit the proposed inference attacks, as done in the case of smashed data.

\begin{table}[t]
	\caption{The model architectures and classification accuracies of learning models. ``FC-3" presents three fully connected layers.}
	\label{tlb:model-baseline}
	\vskip 0.15in
	\begin{center}
		\begin{footnotesize}
			\begin{sc}
				\begin{tabular}{lcccc}
					\toprule
					Data set & \makecell{Bottom\\Model} & \makecell{Top\\Model} & \makecell{Test\\accuracy} \\
					\midrule
					Fashion-MNIST & SimpleNet & FC-3 & 0.9265\\
					CIFAR-10 & SimpleNet & FC-3 & 0.8772\\
					Dogs vs. Cats & SimpleNet & FC-3 & 0.8490\\
					\makecell[l]{Intel Image\\Classification} & SimpleNet & FC-3 & 0.8370\\
					Fruits-360 & SimpleNet & FC-3 & 0.9456\\
					ImageNet & ResNet18 & FC-3 & 0.6158\\
					\bottomrule
				\end{tabular}
			\end{sc}
		\end{footnotesize}
	\end{center}
	\vskip -0.1in
\end{table}

\begin{table*}[ht]
	\caption{The attack performance at the last layer. The attacks based on gradients can achieve $100\%$ accuracy. For attacks from smashed data, the \textit{Euclidean-distance-based inference} outperforms the clustering inference at most times, which is possibly because the clustering characteristics of smashed data are not as salient as the gradients.}
	\label{tlb:attack-perf}
	\vskip 0.15in
	\begin{center}
		\begin{scriptsize}
			\begin{sc}
				\begin{tabular}{*{8}{c}}
					\toprule
					Dataset & \makecell[c]{Number of Classes} & Attack & Training set & Test set\\
					\midrule
					\multirow{8}{*}{Fashion-MNIST} & \multirow{8}{*}{10} & cluster-grad & \textbf{1.000} & \textendash \\
					& & cluster-smashed & 0.924 & \textbf{0.925} \\
					& & Euclid-grad & \textbf{1.000} & \textendash \\
					& & Euclid-smashed & 0.916 & 0.884 \\
					& & Model-completion & 0.487 & 0.487\\
					& & Unsplit & \textbf{1.000} & \textendash \\
					& & LBA & 0.712 & \textendash \\
					& & random-guess & 0.100 & 0.099\\
					\midrule
					\multirow{8}{*}{CIFAR-10} & \multirow{8}{*}{10} & cluster-grad & \textbf{1.000} & \textendash \\
					& & cluster-smashed & 0.797 & 0.737 \\
					& & Euclid-grad & \textbf{1.000} & \textendash \\
					& & Euclid-smashed & 0.833 & \textbf{0.790} \\
					& & Model-completion & 0.261 & 0.262\\
					& & Unsplit & \textbf{1.000} & \textendash \\
					& & LBA & 0.592 & \textendash \\
					& & random-guess & 0.101 & 0.100\\
					\midrule
					\multirow{8}{*}{Dogs vs. Cats} & \multirow{8}{*}{2} & cluster-grad & \textbf{1.000} & \textendash \\
					& & cluster-smashed & 0.980 & 0.833 \\
					& & Euclid-grad & \textbf{1.000} & \textendash \\
					& & Euclid-smashed & 0.975 & \textbf{0.838}\\
					& & Model-completion & 0.501 & 0.495\\
					& & Unsplit & \textbf{1.000} & \textendash \\
					& & LBA & 0.996 & \textendash \\
					& & random-guess & 0.501 & 0.495 \\
					\midrule
					\multirow{8}{*}{Intel Image Classification} & \multirow{8}{*}{6} & cluster-grad & \textbf{1.000} & \textendash \\
					& & cluster-smashed & 0.588 & 0.562 \\
					& & Euclid-grad & \textbf{1.000} & \textendash \\
					& & Euclid-smashed & 0.742 & \textbf{0.710} \\
					& & Model-completion & 0.176 & 0.173 \\
					& & Unsplit & \textbf{1.000} & \textendash \\
					& & LBA & 0.310 & \textendash \\
					& & random-guess & 0.166 & 0.170 \\
					\midrule
					\multirow{8}{*}{Fruits-360} & \multirow{8}{*}{131} & cluster-grad & \textbf{1.000} & \textendash \\
					& & cluster-smashed & 0.950 & 0.911\\
					& & Euclid-grad & \textbf{1.000} & \textendash \\
					& & Euclid-smashed & 0.971 & \textbf{0.935}\\
					& & Model-completion & 0.448 &  0.439 \\
					& & Unsplit & \textbf{1.000} & \textendash \\
					& & LBA & 0.010 & \textendash \\
					& & random-guess & 0.007 &  0.007 \\
					\midrule
					\multirow{8}{*}{ImageNet} & \multirow{8}{*}{1000} & cluster-grad & \textbf{1.000} & \textendash \\
					& & cluster-smashed & 0.367 & 0.197 \\
					& & Euclid-grad & \textbf{1.000} & \textendash \\
					& & Euclid-smashed & 0.374 & \textbf{0.289} \\
					& & Model-completion & 0.211 & 0.164 \\
					& & Unsplit & \textendash & \textendash \\
					& & LBA & \textendash & \textendash \\
					& & random-guess & 0.001 & 0.0006 \\
					\bottomrule
				\end{tabular}
			\end{sc}
		\end{scriptsize}
	\end{center}
	\vskip -0.1in
\end{table*}

\begin{figure}[t]
	\vskip 0.2in
	\begin{center}
		\subfloat[Attack performance on the training set] {
			\begin{minipage}[t]{\linewidth}
				\includegraphics[width=1.0\linewidth]{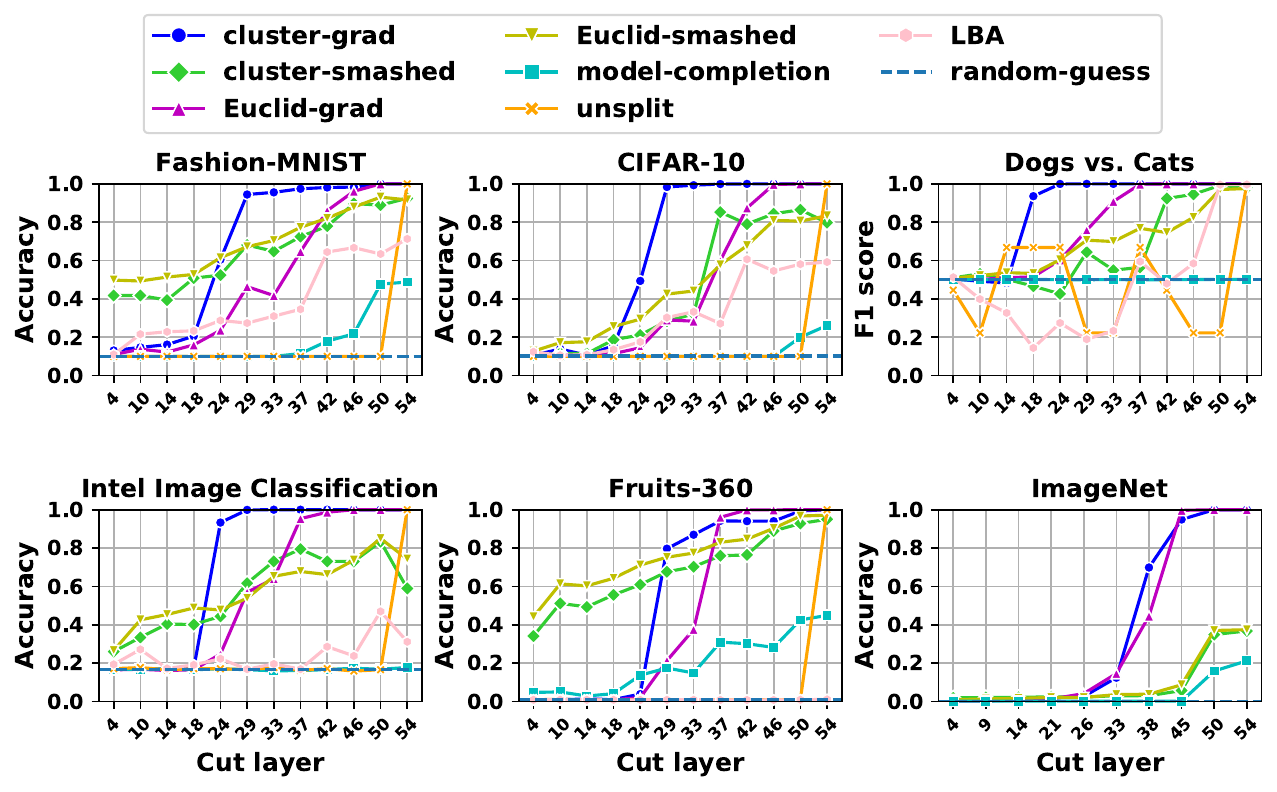}
				\label{fig:layer-train}
			\end{minipage}
		}

		\subfloat[Attack performance on the test set] {
			\begin{minipage}[t]{\linewidth}
				\includegraphics[width=1.0\linewidth]{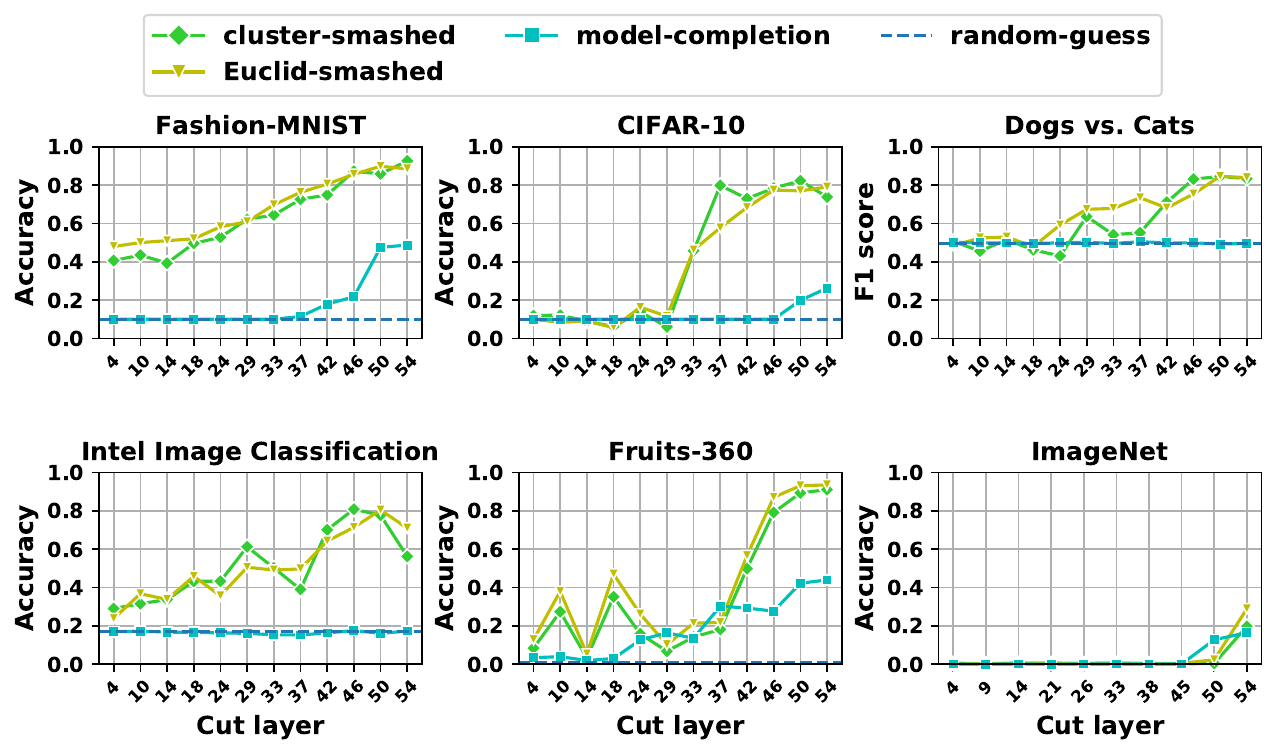}
				\label{fig:layer-test}
			\end{minipage}
		}
		\caption{The attack performance under different positions of cut layers. The attack performance from gradients/smashed data increases with the cut layer approaching to the output.}\label{fig:layer-sensitivity}
	\end{center}
	\vskip -0.2in
\end{figure}

\section{Experiments}\label{sec:experiment}
In this section, we evaluate our proposed label inference attacks via six different datasets for multi-class and binary classification learning tasks.
In the following, we illustrate the experiment environment and datasets in Section~\ref{subsec:exp-setting-up}, and analyze the experimental results of the proposed approach in Section~\ref{subsec:attack-perf}.

\subsection{Datasets and Experiment Environment}
\label{subsec:exp-setting-up}
\textbf{Environment.}
The experimental evaluations are carried out across two distinct workstations.
The first workstation is equipped with Intel(R) Core(TM) i7-10700K CPU @ 3.80GHz, 64GB RAM, and one NVIDIA GeForce RTX 3080 GPU.
In contrast, the second workstation is powered by an 11th Gen Intel(R) Core(TM) i9-11900K CPU @ 3.50GHz, accompanied by 64GB RAM and one NVIDIA GeForce RTX 3090 GPU.
The PyTorch~\cite{Paszke19} learning framework is adopted to implement and execute split learning at the workstation by virtually splitting the DNNs into the bottom and top models.

\textbf{Datasets.} For comprehensive experimental results, we use six different datasets in our experiments, including:
Fashion-MNIST~\cite{xiao17}, CIFAR-10~\cite{Krizhevsky09}, Dogs vs. Cats~\cite{Kaggle13}, Intel Image Classification~\cite{intel18}, Fruits 360~\cite{fruits20}, and ImageNet~\cite{ILSVRC15}.
In particular, for ImageNet, we randomly choose 50 samples per class to evaluate the label inference performance.

\textbf{Models.} We adopt SimpleNet~\cite{simplenet16} and ResNet18~\cite{he2016deep} followed by three fully connected (FC) layers as the classification model to evaluate the datasets.
Table~\ref{tlb:model-baseline} shows the model architectures and performance of the learning models on the classification tasks of six datasets.

\textbf{Baseline and metrics.} 
We conduct a comparative analysis by evaluating the effectiveness of the proposed attacks, compared to random guessing, unsplit attack~\cite{erdogan21}, model completion attack~\cite{Fu22}, and learning-based attack (LBA)~\cite{xie23}.
The effectiveness metrics are the accuracy and F1 scores for multiple and binary classifications, respectively.

Note that unsplit attack~\cite{erdogan21} and LBA~\cite{xie23} involve either exhaustive enumeration of all potential private labels or the optimization of surrogate top model and private labels to align with received gradients.
Such enumeration and optimization would be considerably time-intensive, particularly when applied to extensive datasets with massive label categories.
To this end, we limit our evaluation of these two benchmark techniques to datasets with relatively fewer classes, including Fashion-MNIST, CIFAR-10, Dogs vs. Cats, Intel Image Classification, and Fruits-360. 
This allows us to manage computational complexity while still providing a meaningful assessment of their performance.

For fair comparisons, we randomly select one data sample point per class each time and conduct five times of attacks.
We adopt the average and maximum attack performances as the effectiveness measurements under different layers and defense strategies, respectively.
\subsection{Attack Performance}
\label{subsec:attack-perf}
\begin{figure}[th]
	\vskip 0.2in
	\begin{center}
		\centerline{\includegraphics[width=\columnwidth]{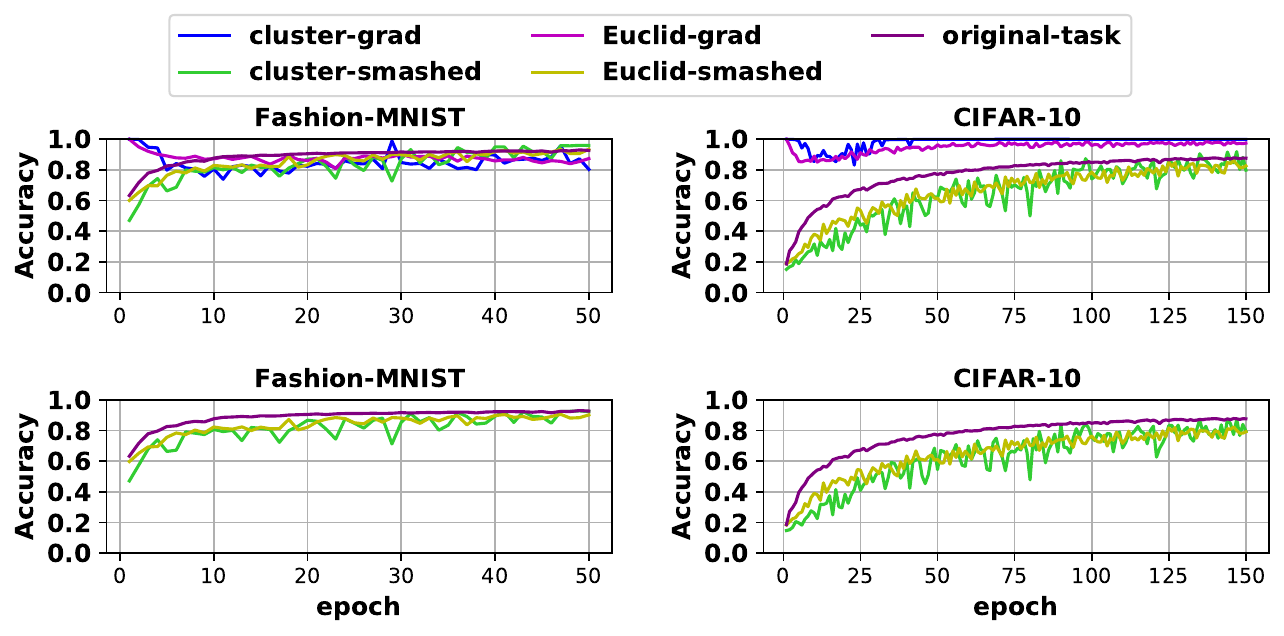}}
		\caption{The attack performance as the gradients/smashed data are collected during different epochs in the training phase. The proposed approach is robust under different epochs. The first and second rows show the attacks on training and test sets, respectively.}
		\label{fig:epoch-sensitivity}
	\end{center}
	\vskip -0.2in
\end{figure}

\begin{figure}[th]
	\vskip 0.2in
	\begin{center}
		\centerline{\includegraphics[width=\columnwidth]{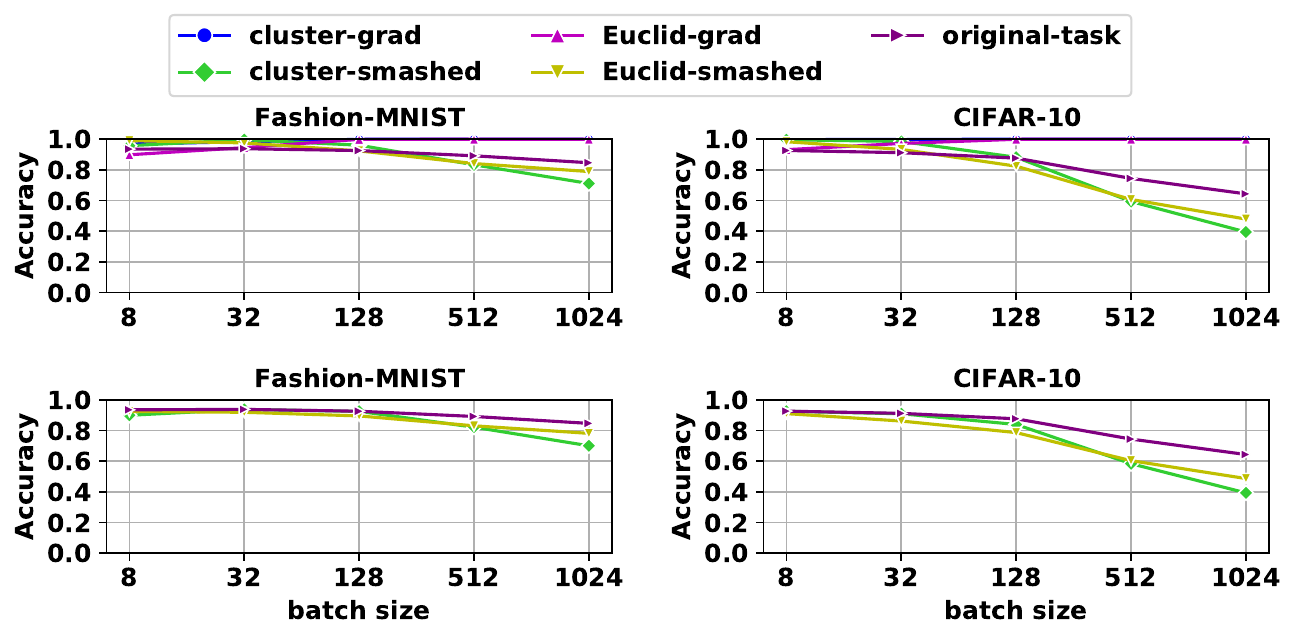}}
		\caption{The attack performance under different batch sizes. The proposed approach is robust to achieve accurate predictions. The first and second row show the attacks on training and test sets, respectively.}
		\label{fig:batch-size-sensitivity}
	\end{center}
	\vskip -0.2in
\end{figure}

\begin{figure}[th]
	\vskip 0.2in
	\begin{center}
		\centerline{\includegraphics[width=\columnwidth]{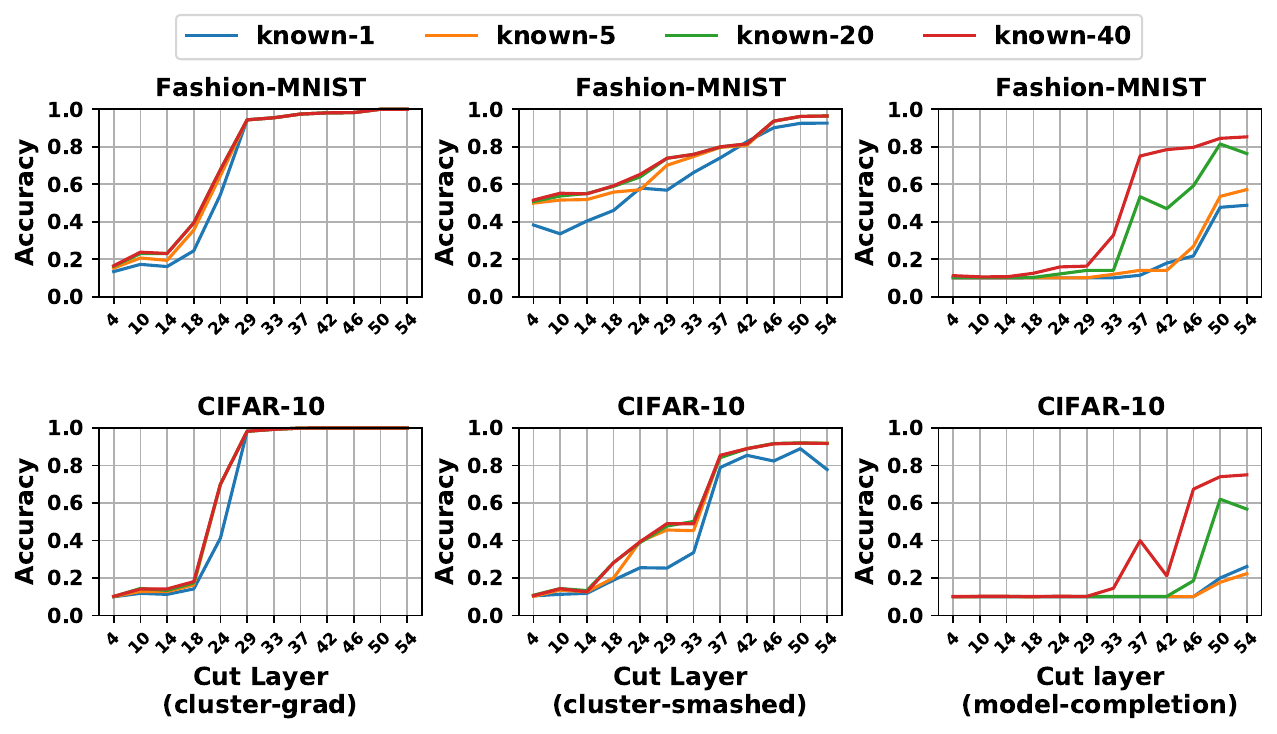}}
		\caption{The attack performance under the different number of known labels. The attack performance is evaluated on the training set of Fashion-MNIST and CIFAR-10.}
		\label{fig:known-quantity}
	\end{center}
	\vskip -0.2in
\end{figure}

\begin{figure}[th]
	\vskip 0.2in
	\begin{center}
		\centerline{\includegraphics[width=\columnwidth]{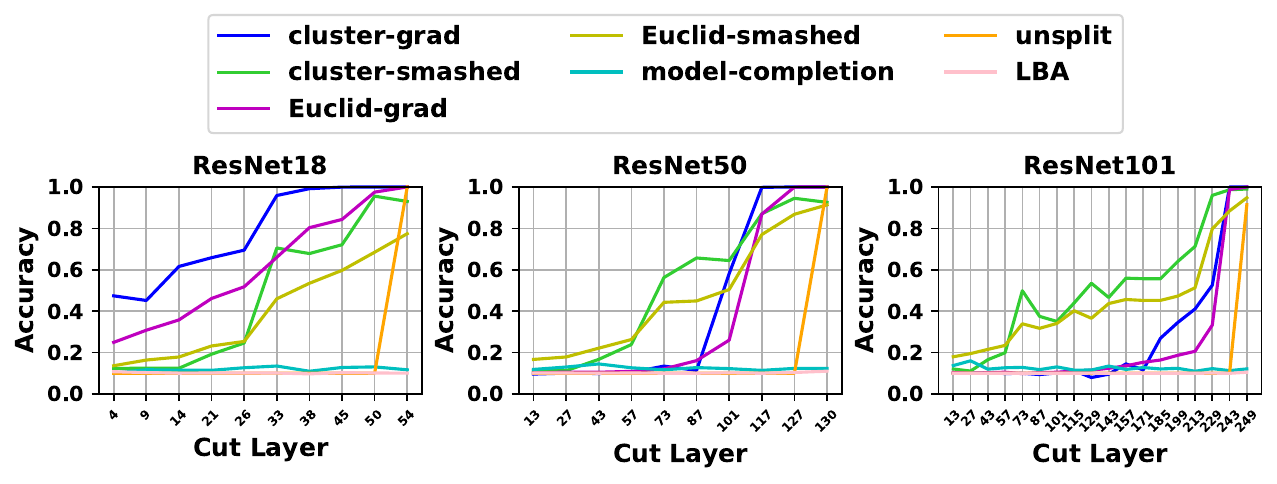}}
		\caption{The attack performance against ResNet-18, RestNet-50, and ResNet-101 on different cut layers.}
		\label{fig:more-layer}
	\end{center}
	\vskip -0.2in
\end{figure}

We evaluate the attack performance of the proposed label inference attacks from the following perspectives.
\begin{itemize}
	\item \textbf{Basic Performance at the Last Layer.} We first analyze the basic performance of the proposed approach at the last layer. The best attack performance is evaluated when the cut layer is close to the output.
	\item \textbf{Different Positions of Cut Layers.} The position of the cut layer has an impact on the attack performance. 
	We show the attack performance under different locations of cut layers.
	\item \textbf{Different Epochs for Attack.}
	The collected gradients and smashed data are changing (vanishing gradients and better-extracted features) with the increase of epochs.
	We evaluate the performance under different epochs to determine the timing for attacks.
	\item \textbf{Robustness under Different Batch Sizes.} 
	It was shown in \cite{zhu19,dang21} that batch size selection in the learning process could degrade the attack performance.
	We aim to validate the robustness of the proposed attacks under different batch sizes.
	\item \textbf{The Number of Labeled Samples for Attacks.} Our proposed attacks only need one labeled sample per class. The attack performance may be related to the number of labeled samples.
	We evaluate the performance under different amounts of labeled samples per class.
	\item \textbf{Attack Performance of Transfer Learning.} For attacks based on smashed data, the attacker can exploit the pre-trained models of similar tasks.
	We evaluate the performance under different models.
	\item \textbf{Attack Performance against the Increasing Model Layers.}
	The attack performance may be influenced by the increase of the model layers.
	Thus, we evaluate the attack performance against ResNet-18, ResNet-50, and ResNet-101, respectively.
\end{itemize}

\textbf{1) Basic Performance at the Last Layer.}
Table~\ref{tlb:attack-perf} summarizes the attack performance of the proposed approaches from the gradients/smashed data at the last layer.
We can see that the attack based on the collected gradients can reconstruct all the private labels correctly for all the datasets, even for the most sophisticated dataset \textit{ImageNet}, which has 1000 classes. 

The attacks based on the smashed data can also achieve good performance.
The \textit{Euclidean-distance-based} method usually achieves better performance than the clustering-based method, which is possibly because the clustering characteristics of smashed data are not as salient as the gradients.
From Table~\ref{tlb:attack-perf}, we can observe that the attacks based on the smashed data perform better on the training set than the test set, as the final model is trained on the training set. 
Based on the smashed data, the proposed attacks can achieve great performance with an accuracy/F1 score of around 0.95 on training sets of Fashion-MNIST, Dogs vs. Cats, and Fruits-360. 
For the test sets of Fashion-MNIST and Fruits-360, the attack performance can reach about 0.92. On CIFAR-10, our proposed attacks based on smashed data achieve an accuracy of around 0.80, while around 0.72 on Intel Image Classification. 
For the most complicated ImageNet, the attack accuracy based on smashed data is around 0.30. 
Considering that the test accuracy of the trained model is 0.6158 and the random guess can only achieve around 0.001, the attack performance is not bad.

For the model completion attack on Dogs vs. Cats and Intel Image Classification, the attack performance degrades to the random guess. 
The reason is possibly that the datasets with limited classes can only provide limited labeled samples (one labeled sample per class), which is not enough for the model completion attack to train an effective surrogate top model.
Unsplit attack would be efficient when the cut layer is at the last layer. 
However, the label enumeration in unsplit attack also faces the scalability problem and can only operate in the case of stochastic gradient descent (i.e., setting the mini-batch size to one).
For simple datasets, e.g., Fashion-MNIST, CIFAR-10 and Dogs vs. Cats, LBA can achieve effective attack performance.
However, LBA cannot infer the private labels of sophisticated datasets effectively.
The high-dimensional gradients and the large number of classes make the optimization hard to work accurately.
Note that our proposed attacks can outperform the model completion, unsplit and LBA on all the datasets.

\textbf{2) Different Positions of Cut Layers.} 
Fig.~\ref{fig:layer-sensitivity} plots the attack accuracies from the gradients/smashed data under different positions of cut layers.
We can see in Fig.~\ref{fig:layer-sensitivity} that the attack performance from gradients/smashed data improves with the cut layer approaching the output layer.
In Fig.~\ref{fig:layer-sensitivity}, even if the cut layer is far from the output layer, the attack performance based on collected gradients can still achieve accurate inference. 
The attack performance of the \textit{clustering-based} method exhibits more robustness to the cut position than the \textit{Euclidean-distance-based} method for the attacks based on gradients.
Besides, the gradients-based inferences on the datasets with fewer classes are more robust to the position of the cut layer.
For example, when the cut layer is located at the 18th layer, the attack performance based on gradients can achieve an accuracy of about 0.90 for Dogs vs. Cats that only has two classes, while the attack accuracies on other datasets degrade to the random guess.
Except the batch size constraint, we can observe that unsplit attack can infer the private labels with high accuracy only when the cut layer is just before the output layer.
Once the cut layer moves towards the input, unsplit cannot obtain the private labels accurately.
In Fig.~\ref{fig:layer-sensitivity}\subref{fig:layer-train}, we can see that when the cut layer gets close to the input layer, F1 scores of unsplit and LBA fluctuate around 0.5 on Dogs vs. Cats.
This is because the sophisticated top models make the gradient matching noneffective.

\textbf{3) Different Epochs for Attack.} 
Fig.~\ref{fig:epoch-sensitivity} plots the attack accuracies as the gradients/smashed data are collected during different epochs in the training process.
Again, the attack performance from gradients achieves close to $1.0$ accuracy. The attack performance from smashed data performs increasingly well with the number of epochs.
This is because the learning models are trained close to convergence and can better extract the features (suitable for attack from smashed data).
We also note that the attack performance from gradients fluctuates after several epochs, possibly because the gradients of activations in different batches fluctuate greater after the initial training.
From Fig.~\ref{fig:epoch-sensitivity}, we can conclude that the inference attack from gradients should be better executed \textit{at the beginning of the training process}.
The attack from smashed data may be suitable \textit{after the training process}.

\textbf{4) Robustness under Different Batch Sizes.} Fig.~\ref{fig:batch-size-sensitivity} plots the attack performance under different batch sizes.
We can see that the proposed attack approach is robust and consistently achieves accurate predictions from the gradients under different batch sizes.
For the attacks based on the smashed data, the attack performance decreases with the increase of the batch size.
That is because a large batch size slows down the training and the model does not converge under the same epochs.

\textbf{5) The number of Labeled Samples for Attack.} Fig.~\ref{fig:known-quantity} plots the attack performance under different amounts of labeled samples per class.
We evaluate the attack accuracy on the training set with our clustering method and model completion attack. 
The attack accuracy grows with the increase of the labeled samples per class. Model completion attack hardly increases the attack performance when the number of the known samples per class reaches 40~\cite{Fu22}.
We can see that for an effective model completion attack, the attacker needs much more labeled samples than our proposed attacks. Even with enough labeled samples, our proposed attacks outperform the model completion attack in terms of the scalability of cut layer locations.

\textbf{6) Attack Performance of Transfer Learning.} Table.~\ref{tlb:transfer-accuracy} reports the attack performance of the Euclid distance based attack on CIFAR-10, Dogs vs. Cats, Intel Image Classification, and Fruits-360 under different pre-trained models, including AlexNet, VGG16, ResNet18 and ResNet50.
As shown in Fig.~\ref{fig:layer-sensitivity}, the attack performance via gradients and smashed data may be vulnerable to the cut layer locations, especially when the cut layer is close to the input layer. 
To this end, the adversary can extract the features using the published learning models of similar tasks.
The attack performance does not depend on the cut layer locations, and the attack accuracy can be up to 0.846 in ResNet50 on Dogs vs. Cats dataset.

\textbf{7) Attack Performance against the Increasing Model Layers.}
Fig.~\ref{fig:more-layer} shows the attack performance against ResNet-18, RestNet-50, and ResNet-101.
We can see that the gradient-based attack performance degrades with the increasing model layers, and the attacks based on the smashed data can get better attack performance.
This is because the increase of model layers can help better extract the discriminative features of the data input. 
As a result, the smashed data would become increasingly effective with the increasing number of layers. 
In contrast, the one-layer gradient vector would carry less information with the increasing depth of a learning model, leading to the degradation of the attack performance.

\begin{table}[!t]
	\caption{The attack performance of transfer learning.}
	\label{tlb:transfer-accuracy}
	\vskip 0.05in
	\begin{center}
		\begin{footnotesize}
			\begin{sc}
				\setlength{\tabcolsep}{1mm}
				\begin{tabular}{lccccc}
					\toprule
					Dataset & AlexNet & VGG16 & Resnet18 & Resnet50 \\
					\midrule
					CIFAR-10 & 0.4678 & 0.4430 & \textbf{0.5781} & 0.5150\\
					Dogs vs. Cats & 0.5571 & 0.8441 & 0.6560 & \textbf{0.8460}\\
					\makecell[l]{Intel Image\\Classification} & 0.6138 & 0.6558 & 0.6664 & \textbf{0.7383}\\
					Fruits-360 & \textbf{0.6116} & 0.5789 & 0.5755 & 0.6072\\
					\bottomrule
				\end{tabular}
			\end{sc}
		\end{footnotesize}
	\end{center}
	\vskip -0.3in
\end{table}

\begin{figure}[t]
	\vskip 0.2in
	\begin{center}
		\centerline{\includegraphics[width=\columnwidth]{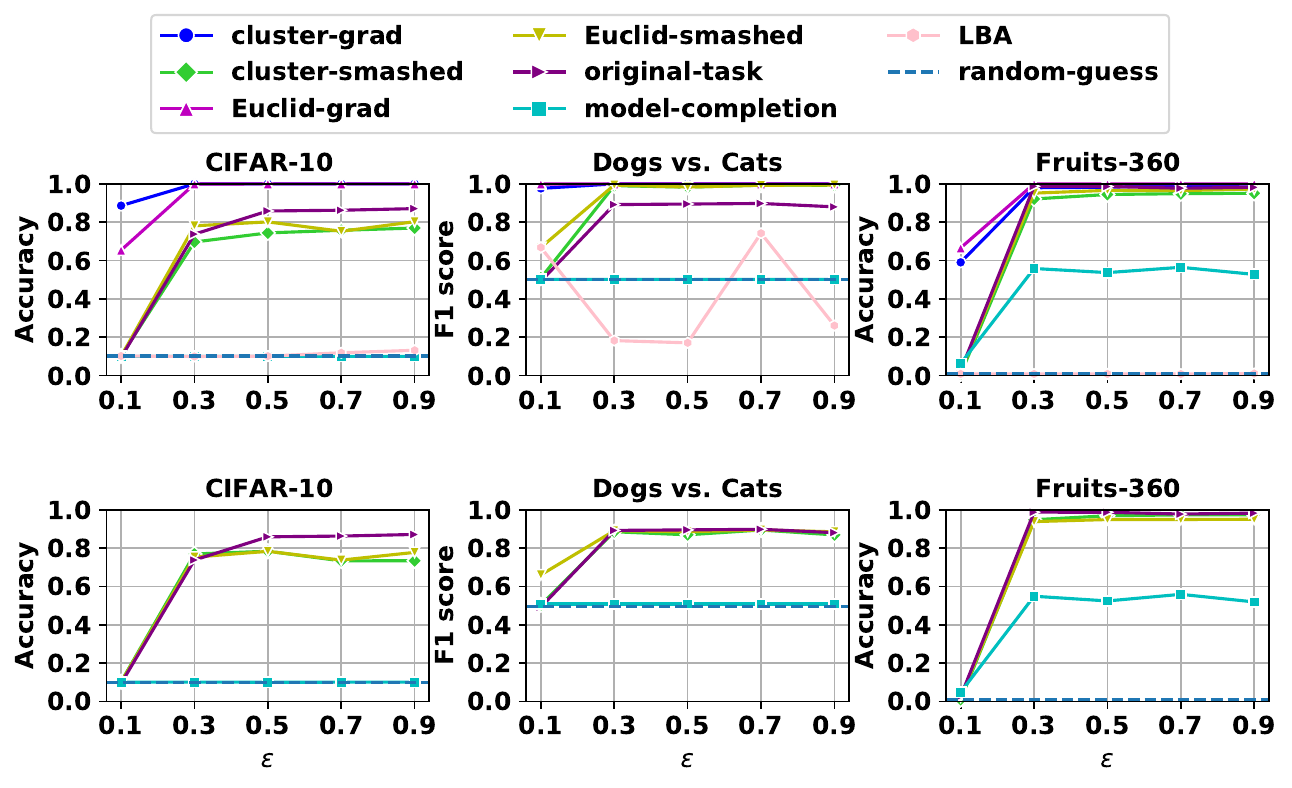}}
		\caption{The attack performance against DP-SGD, where the coefficient $\epsilon$ ranges from 0.1 to 0.9. DP-SGD cannot protect the labels in the proposed approach.}
		\label{fig:dp-sensitivity}
	\end{center}
	\vskip -0.2in
\end{figure}

\begin{figure}[ht]
	\vskip 0.2in
	\begin{center}
		\centerline{\includegraphics[width=\columnwidth]{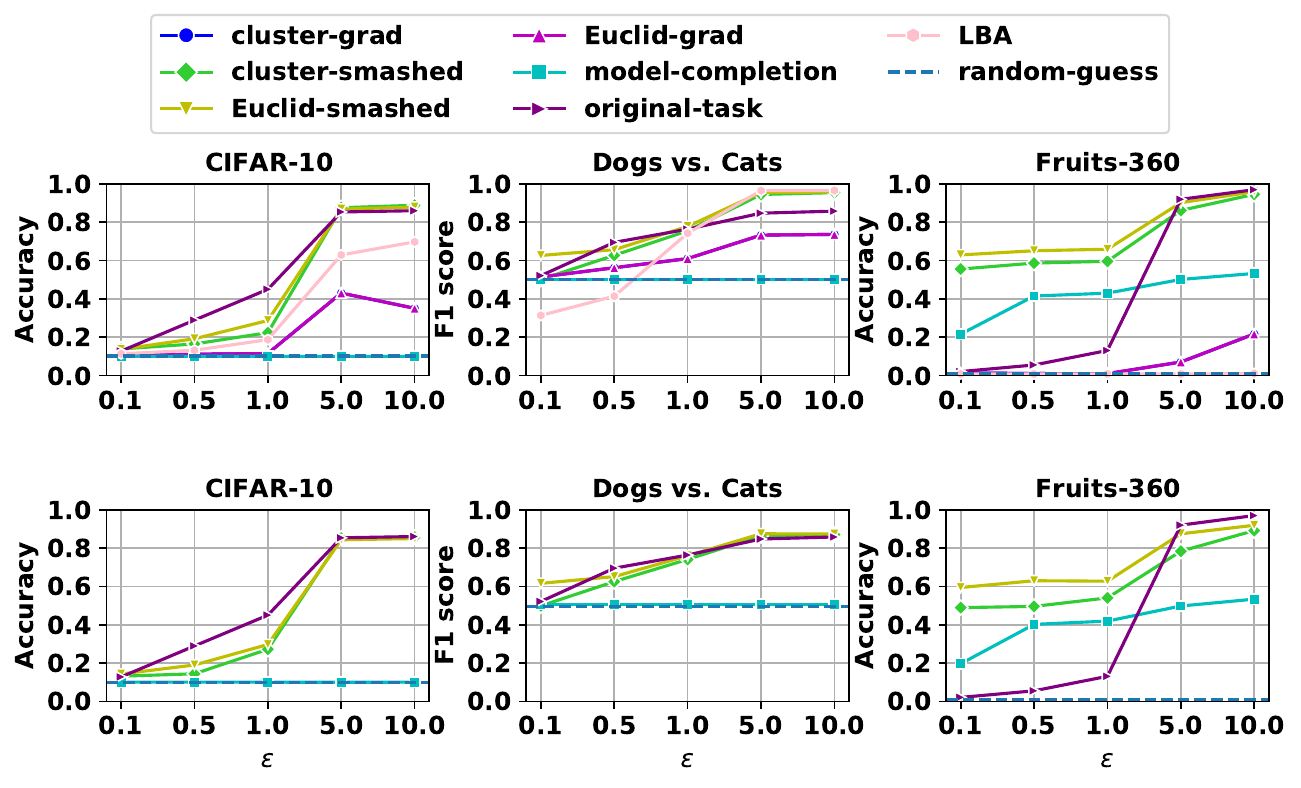}}
		\caption{The attack performance against label differential privacy. Label differential privacy can mitigate the attacks based on gradients, but exhibit bad defense performance against the attacks based on smashed data.}
		\label{fig:label-dp}
	\end{center}
	\vskip -0.2in
\end{figure}

\begin{figure}[ht]
	\vskip 0.2in
	\begin{center}
		\centerline{\includegraphics[width=\columnwidth]{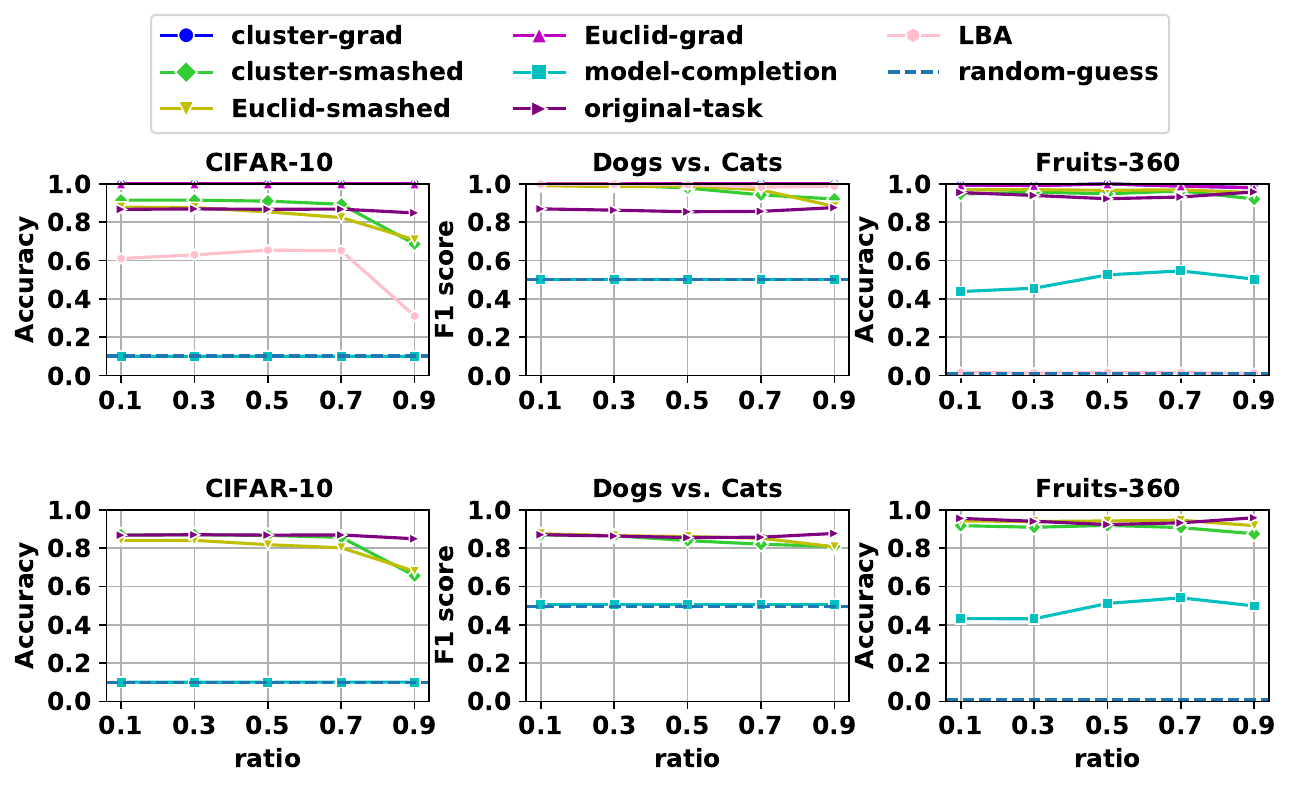}}
		\caption{The attack performance against gradient compression. The proposed attack from gradients can consistently achieve accurate predictions, regardless of the compression ratios.}
		\label{fig:grad-pruning-sensitivity}
	\end{center}
	\vskip -0.2in
\end{figure}

\begin{figure}[ht]
	\vskip 0.2in
	\begin{center}
		\centerline{\includegraphics[width=\columnwidth]{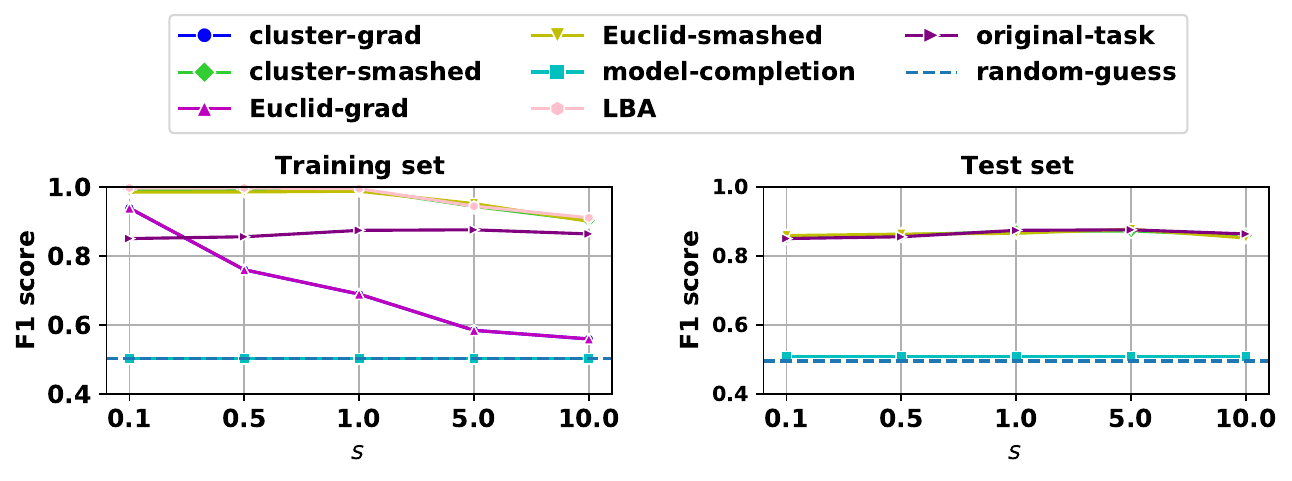}}
		\caption{The attack performance against Marvell. Marvell fail to mitigate the attacks based on smashed data.}
		\label{fig:marvell}
	\end{center}
	\vskip -0.2in
\end{figure}
\section{Attack against Defenses}
As shown in Section~\ref{sec:experiment}, the proposed label inference attacks from gradients can achieve accurate and robust predictions under different settings.
This section evaluates the attack performance of the proposed approach against the existing defense techniques.
In particular, we select four state-of-the-art defense techniques, including 1) DP-SGD~\cite{Abadi16}, 2) label differential privacy (LDP)~\cite{Ghazi21}, 3) gradient compression (i.e pruning the gradients with small magnitudes)~\cite{tsuzuku18, lin20}, and 4) Marvell~\cite{li22}.

For fair comparisons, the cut layer is designed to be located before the fully connected layers (i.e., the last three layers), as done in~\cite{Fu22}.
As unsplit atrtack only works at the last layer, we exclude unsplit attack from our defense evaluation.
From the experimental results in Section~\ref{sec:experiment}, we select three datasets (i.e., CIFAR-10, Dogs vs. Cats, and Fruits-360) to evaluate the attack performance against the defenses.

\textbf{1) DP-SGD.}
Differential privacy is a popular defense technique that meticulously designs the artificial noise for the data points (e.g., gradients) to provide a theoretical privacy guarantee (see the following definition).

\begin{definition}
	\label{def:dp}
	$(\epsilon, \delta)$-DP: An algorithm $\mathcal{M}: \mathcal{D} \rightarrow \mathcal{R}$ is $(\epsilon, \delta)$-differential private if for any two datasets $D$, $D^\prime$ which differ on at most one row, and every set of outputs $\mathcal{O} \in \mathcal{R}$:
	\begin{equation}
		\mathcal{P}\left[\mathcal{M}(D) \in \mathcal{O}\right] \leq e^\epsilon \mathcal{P}\left[\mathcal{M}(D^\prime) \in \mathcal{O}\right] + \delta\text{.}
	\end{equation}
\end{definition}

Consider the setting of the label inference attack.
The victim (label party) can train the top model with differential privacy~\cite{dwork14} to prevent the label inference attack from the adversary.
DP-SGD~\cite{Abadi16} algorithm is adopted to generate the artificial noise to the gradients via Opacus~\cite{Yousefpou21}.
The two control parameters in DP are set as follows:
1) $\delta$ is designed to be the inverse of the size of the training dataset, and 2) $\epsilon$ is the adjustable variable in the simulations (where small values of $\epsilon$ indicate large noise and high privacy guarantee).

Fig.~\ref{fig:dp-sensitivity} plots the attack performance against differential privacy.
We can see that differential privacy can successfully mitigate LBA.
The injected noise in gradients prevents the adversary to infer private labels through gradient matching.
However, we can see that differential privacy cannot protect the private labels against our proposed attacks.
In other words, the inaccurate label inference based on the smashed data (e.g., $\epsilon < 0.3$) is at the cost of inaccurate performance of the learning task.
This is because the gradients are based on the loss function (with the correct label) and carry more information than the forward inference.
The large levels of artificial noises can quickly destroy the accuracy of the forward inference for learning models.

\textbf{2) Label Differential Privacy.}
Label differential privacy (LDP)~\cite{Ghazi21} is proposed to protect label privacy during training, which injects noisy labels to mask the ground-truth labels.
LDP algorithm adopts a randomized response mechanism to mask the ground truth label. $\epsilon$ is the adjustable variable in the simulations (where small values of $\epsilon$ indicate a high privacy guarantee). LDP needs the prior distribution of labels. 
Without loss of generality, we directly calculate the label distribution from the dataset.

Fig.~\ref{fig:label-dp} plots the attack performance against label differential privacy.
We can see that the attack accuracy of gradients degrades under label differential privacy. 
However, label differential privacy cannot protect private labels from attacks in terms of the smashed data.
The inaccurate label inference (e.g., when $\epsilon < 5$ in CIFAR-10, Dogs vs. Cats, and Fruits-360) is at the cost of inaccurate performance of the learning task.
For simple datasets, e.g., CIFAR-10 and Dogs vs. Cats, LDP cannot provide effective defense performance, either.
A few noisy labels cannot prevent LBA from optimizing the private labels by gradient matching.

\textbf{3) Gradient Compression.} 
Gradient compression~\cite{tsuzuku18, lin20} is widely adopted to speed up the training process and protect data privacy.
In gradient compression, the gradients with small magnitudes are pruned.
The control variable is the gradient compression ratio, i.e., the percentage of pruned gradients.

Fig.~\ref{fig:grad-pruning-sensitivity} shows the attack performance under different gradient compress ratios varying from $0.1$ to $0.9$. 
We can see that the compression of gradients during training influences the quality of the learning model, resulting in a slight decrease in classification accuracy under large compression ratios.
The proposed attacks from gradients can consistently achieve accurate predictions of private labels, regardless of the compression ratios.
This validates the robustness of the proposed approach against gradient compression.
When the compression ratio is configured to 0.9, the application of gradient compression demonstrates its effectiveness in mitigating LBA on CIFAR-10 dataset.
The reason may be that the compressed cut-layer gradients cannot provide enough information for optimization.
However, for binary classification, LBA can still achieve a good attack performance despite a compression ratio of 0.9.
It demonstrates that LBA exhibits more robustness on the datasets with fewer classes.

\textbf{4) Marvell.} 
Marvell is designed for binary classification and injects noise into the gradients through optimization. The privacy parameter $s$ controls the noise intensity. The larger $s$ means more significant noise.

Fig.~\ref{fig:marvell} shows the attack performance under different $s$ from $0.1$ to $10$. 
We can see that the typical values of $s \in \{0.1, 0.5, 1\}$ cannot provide efficient protection against the proposed label inference attack, especially for the attacks from smashed data. 
The large value of $s \in \{5.0, 10.0\}$ can protect the private labels against attacks from gradients but fail to defend the attacks based on smashed data and LBA.
\textbf{Summary.} The experiments show that DP-SGD, LDP, Gradient Compression, and Marvell cannot provide comprehensive protection against our proposed attacks.
	The possible reasons may be as follows.
	\begin{itemize}
    \item Some existing defense mechanisms (i.e., label differential privacy and Marvell) were mainly designed to protect the information leakage from the exchanged gradients (e.g., by adding noises to the label). 
    The proposed approaches defeat these defenses by directly inferring the labels from the exchanged smashed data (i.e., embedding).
    \item DP-SGD is typically used to anonymize the identities of the batch samples through gradient perturbation in federated learning.
    However, the non-label party (adversary) knows the sample identities in the design of split learning.
    DP-SGD cannot provide effective protection in split learning.
    \item Gradient compression is designed to prune the gradients with small magnitudes to protect against the attacks requiring accurate gradient information, e.g., deep leakage attack~\cite{zhu19}. 
    However, our proposed attacks are based on the gradient or embedding similarity.
    The pruned gradients with small magnitudes failed to change the key features of the similarities, thus failing to defeat the proposed attack.
  	\end{itemize}
	A potential countermeasure should consider the leakage from gradients and extracted embeddings comprehensively.
	For example, we can find or design some criteria to measure the label leakage from the gradients and embeddings.
  	By jointly minimizing the label leakage measure (e.g., adding as the penalty terms in the training process), we can jointly protect both the gradients and embeddings.
  	One possible criterion to measure the leakage from the embeddings is the correlation between the cut-layer embeddings and the private labels.
\section{Conclusion}
This paper designs three label inference attacks against split learning based on the unified similarity measurement. The adversary can perform passive attacks from gradients and smashed data during both the training and inference phases.
We mathematically analyze four possible label leakages and propose the cosine and Euclidean similarity measurements for gradients and smashed data, respectively.
Validated from the experiments, the proposed attack from gradients can achieve close to $100\%$ accuracy for split learning under various settings.
The proposed attack is also robust to the state-of-the-art defense techniques, which motivates future works on new efficient defense mechanisms for split learning.
Moreover, in our future work, we will conduct an in-depth investigation into the broader applicability of the proposed similarity measurement technique, including but not limited to membership attacks.

\

\bibliographystyle{IEEEtran}
\bibliography{bare_jrnl_new_sample4}

\vfill

\end{document}